\documentclass{article} % For LaTeX2e
\usepackage[preprint]{colm2025_conference}

\usepackage{microtype}
\usepackage{hyperref}
\usepackage{url}
\usepackage{booktabs}
\usepackage{graphicx}
\usepackage{colortbl}
\usepackage{tabularx} 
\usepackage{array}
\usepackage{csquotes}
\usepackage{lineno}

\definecolor{darkblue}{rgb}{0, 0, 0.5}
\hypersetup{colorlinks=true, citecolor=darkblue, linkcolor=darkblue, urlcolor=darkblue}

\usepackage{subcaption}
\usepackage{wrapfig}
\usepackage{amsmath,amsfonts,bm}
\usepackage{xspace}
\usepackage{fancyvrb}
\usepackage{fvextra} % Extended features for fancyvrb
\usepackage[most]{tcolorbox}

\lstdefinestyle{mystyle}{
    columns=fullflexible,
    backgroundcolor=\color{backcolour},   
    commentstyle=\color{codegreen},
    numberstyle=\tiny\color{codegray},
    stringstyle=\color{codepurple},
    basicstyle=\footnotesize,
    breakatwhitespace=false,         
    breaklines=true,  
    breakindent=0pt,
    captionpos=b,                    
    keepspaces=true,  
    numbersep=0pt,                  
    showspaces=false,                
    showstringspaces=false,
    showtabs=true,                  
    tabsize=10,
    escapeinside={@}{@}, % Define escape sequence for LaTeX inside listings
    moredelim=**[is][\highlight]{`}{`}, % Highlight words with `
}
\everypar{\looseness=-1}
\usepackage{titlesec}

\titlespacing*{\section}
{0pt}{1.9ex plus 1ex minus .5ex}{.7ex plus .4ex}

\titlespacing*{\subsection}
{0pt}{1ex plus 1ex minus .5ex}{.5ex plus .4ex}

\title{
LongPerceptualThoughts:\\ Distilling System-2 Reasoning  for System-1 Perception
}

\author{
  \textbf{Yuan-Hong Liao}${^{\spadesuit}}$
  \and
  \textbf{Sven Elflein}${^{\spadesuit\diamondsuit}}$
  \and
  \textbf{Liu He}${^{\clubsuit}}$ 
  \and
  \textbf{Laura Leal-Taixé}${^{\diamondsuit}}$ 
  \AND
  \textbf{Yejin Choi} ${^{\diamondsuit}}$
  \and
  \textbf{Sanja Fidler}${^{\spadesuit\diamondsuit}}$
  \and
  \textbf{David Acuna}${^{\diamondsuit}}$
}

\newcommand{\onedot}{.\xspace}

\def\eg{\emph{e.g}\onedot} 

\def\ie{\emph{i.e}\onedot}

\def\etc{\emph{etc}\onedot}

\newcommand{\AL}[1]{{\color{brown}{[Andrew: #1]}}}
\newcommand{\DA}[1]{{\color{red}{[David: #1]}}}

\newcommand{\yejin}[1]{{\color{cyan}{[Yejin: #1]}}}

\newcommand{\longcot}{LongPerceptualThoughts\xspace}

\newcommand{\model}{\mathcal{M}}           % Reasoning model
\newcommand{\VLM}{\mathcal{M}_\text{VLM}}
\newcommand{\LLM}{\mathcal{M}_\text{LLM}}
\newcommand{\RLLM}{\mathcal{M}_\text{Reason}}
\newcommand{\question}{\rq}                  % Input question
\newcommand{\image}{\rv}
\newcommand{\longthought}{\ermZ}
\newcommand{\stepthought}{\rz}

\newcommand{\reasoning}{\rz}                 % Full reasoning trace 
\newcommand{\densecaption}{\rc}
\newcommand{\step}{\rs}

\newcommand{\marker}{{\textnormal{m}}}              % Single step
\newcommand{\thought}[1]{\tau_{#1}}           % Single thought (as a collection of steps)
\newcommand{\newreasoning}{\bar{z}}                 % Full reasoning trace 

\newcommand{\answer}{\ra}                    % Final answer
\newcommand{\basemodel}{\texttt{BaseModel}\xspace}
\def\eqref#1{equation~\ref{#1}}
\def\1{\bm{1}}

\def\ra{{\textnormal{a}}}

\def\rc{{\textnormal{c}}}

\def\rq{{\textnormal{q}}}

\def\rs{{\textnormal{s}}}

\def\rv{{\textnormal{v}}}

\def\rx{{\textnormal{x}}}

\def\rz{{\textnormal{z}}}

\def\ermZ{{\textnormal{Z}}}

\DeclareMathAlphabet{\mathsfit}{\encodingdefault}{\sfdefault}{m}{sl}
\SetMathAlphabet{\mathsfit}{bold}{\encodingdefault}{\sfdefault}{bx}{n}

\begin{document}

\ifcolmsubmission
\linenumbers
\fi

\maketitle

\begin{abstract}

Recent reasoning models through test-time scaling have demonstrated that long chain-of-thoughts can unlock substantial performance boosts in hard reasoning tasks such as math and code. However, the benefit of such long thoughts for system-2 reasoning is relatively less explored in other domains such as perceptual tasks where shallower, system-1 reasoning seems sufficient.
In this paper, we introduce \textit{\longcot}, a new synthetic dataset with 30K long-thought traces for perceptual tasks.
The key challenges in synthesizing elaborate reasoning thoughts for perceptual tasks are that off-the-shelf models are not yet equipped with such thinking behavior and that it is not straightforward to build a reliable process verifier for perceptual tasks. Thus, we propose a novel three-stage data synthesis framework that first synthesizes verifiable multiple-choice questions from dense image descriptions, then extracts simple CoTs from VLMs for those verifiable problems, and finally expands those simple thoughts to elaborate long thoughts via frontier reasoning models.
In controlled experiments with a strong instruction-tuned 7B model, we demonstrate notable improvements over existing visual reasoning data-generation methods. 
Our model, trained on the generated dataset, achieves an average +3.4 points improvement over 5 vision-centric benchmarks, including +11.8 points on V$^*$ Bench. 
Notably, despite being tuned for vision tasks, it also improves performance on the  text reasoning benchmark, MMLU-Pro, by +2 points.
\footnote{Project website: \url{https://andrewliao11.github.io/LongPerceptualThoughts}}
%Furthermore, we find the responses of the model are not only longer, but also exhibit characteristic system-2 behaviors, highlighting the effectiveness of our data.
% in promoting richer test-time scaling.
% We will release a dataset of 50K samples tailored for both supervised fine-tuning and preference-based reinforcement learning, providing a valuable resource for the broader multimodal reasoning community.
% We propose a new algorithm to generate multimodal reasoning data by injecting key cognitive behaviors distilled from reasoning LLMs (e.g., R1) in current VLM's response to existing visual datasets (say captioning/multiple choice)
%DA: add the curriculum stylization part to stay in distribution.
%
% Our pipeline works for both SFT and RL.
% Together with the paper, we also plan to release a 50K dataset for both SFT and RL (preference)

\end{abstract}

\renewcommand{\thefootnote}{\fnsymbol{footnote}}
\footnotetext[0]{\hspace{-4pt}\textsuperscript{$^\spadesuit$}University of Toronto, Vector Institute, \textsuperscript{$^\diamondsuit$}NVIDIA, \textsuperscript{$^\clubsuit$}Purdue University}
\renewcommand{\thefootnote}{\arabic{footnote}}

\section{Introduction}\label{sec:intro}

% \AL{One thing that I think we can highlight is that we are not doing math problem. I think this is important in VLMs. Most reasoning LLMs foundings are in specifically in math problems. In this work, we are tackling a broader problems. Additionally, our data generation pipeline does not have strict constraints.}

Reasoning models, such as OpenAI’s o1~\citep{openai2024_o1} and Deepseek’s R1~\citep{deepseekai2025deepseekr1incentivizingreasoningcapability}, have demonstrated remarkable capabilities in solving complex reasoning problems by scaling test-time compute. Intuitively, they increase the number of tokens generated at inference-time, allowing the model to ``think longer''—producing longer chain-of-thoughts (CoTs)  that go beyond typical linear rationales that mimic textbook responses.

\iffalse
In practice, these extended reasoning chains can be viewed as an implicit internal search mechanism, wherein the model implicitly explores multiple potential solution paths, verifies intermediate conclusions, and self-corrects when necessary—cognitive behaviors strikingly similar to human problem-solving \citep{gandhi2025cognitivebehaviorsenableselfimproving,xiang20252reasoningllmslearning}.
\fi

\iffalse
The success of test-time scaling in commercial reasoning models has sparked significant interest among the academic community aiming to replicate these results through several techniques~\citep{xx}. 
% such as searching~\citep{}, multi-stage reinforcement learning training~\citep{}, and 
% distillation. 
In particular, many open-source projects aim to replicate these capabilities by training on  long CoT traces distilled from frontier reasoning models \citep{sky_t1_2025,openthoughts}.
These methods have demonstrated cost-efficient performance reproduction in domains such as math, code, and other verifiable textual tasks.
\fi
However, despite numerous attempts to match the performance of models like o1 or R1 on challenging math benchmarks~\citep{AIME2024,lightman2023let}, less effort has been directed toward tasks beyond mathematical reasoning. 
% Most notably, how to cost-effectively generate synthetic, long-form reasoning traces for \textit{vision-centric reasoning tasks}, suitable for both supervised fine-tuning and preference-based RL, remains largely underexplored. 
% \AL{I think we need to explicitly associate reasoning with the term ``CoT''.}
Most notably, how to 
% cost-effectively 
generate synthetic, long-form CoT reasoning traces that solve \textit{vision-centric tasks}—suitable for both supervised fine-tuning and preference-based RL—remains largely underexplored. 
% \AL{should be make the term consistent? vision-centric tasks vs. vision-centric reasoning tasks vs. perceptual tasks}\DA{agree. and maybe somewhere in preliminary have a small sentence or two defining it. they may not be familiar for the reader} 
% \AL{I vote for vision-centric tasks and we approach them via extended reasoning.}
% \DA{add that this could also be achieved only with  RL}
% \DA{should we add RL alone could lead to this behaviours, or maybe is too much in VLMs?}

Vision-centric tasks have proven challenging for
% for both open-source and proprietary 
vision-language models (VLMs), especially when the tasks require 
% cases requiring visual commonsense, 
object counting and localization, scene understanding, and 2D/3D spatial reasoning~\citep{liao-etal-2024-reasoning,Rahmanzadehgervi_2024_ACCV,campbell2024understanding}. 
Prior works have addressed these challenges by helping VLMs ``see'' better. Common approaches include modifying the input image (\eg, through cropping)
% generating intermediate visual artifacts (\eg drawings), 
or incorporating intermediate representations,  into the CoT~\citep{v_star,visual_cot,wu2025groundedchainofthoughtmultimodallarge}. 
In contrast, we propose to synthesize data that implicitly equips VLMs with an internal search mechanism—one that unfolds through a textual inner monologue, enabling the model to explore multiple potential solution paths: revisiting  different image regions, verifying  intermediate conclusions, identifying  inconsistencies, and self-correcting  when necessary. 
% 
% As shown in Fig.~\ref{fig:teaser}, these human-like reasoning behaviors effectively covers a diverse set of solution spaces.
% 
Our approach is complementary to prior methods and mirrors the behavior observed in reasoning models like R1 and commercial VLMs such as o1, where reasoning performance improves by scaling test-time inference.
We emphasize we do not claim that long textual CoTs are inherently superior or the only way to scale test-time inference in VLMs. Rather, our goal is to synthesize data to equip models with such capability—an approach shown to be effective in SoTA reasoning models. Furthermore, given the difficulty of building reliable process verifiers for search in perceptual tasks, our data-centric method offers a practical alternative.

% —\ie letting the model ``think longer'' via extended CoTs. 

\begin{figure}[t]
  \centering
  % Placeholder box (width=6cm, height=4cm)
  %\fbox{\rule{0pt}{8cm} \rule{\textwidth}{0pt}} 
  \includegraphics[width=\textwidth]{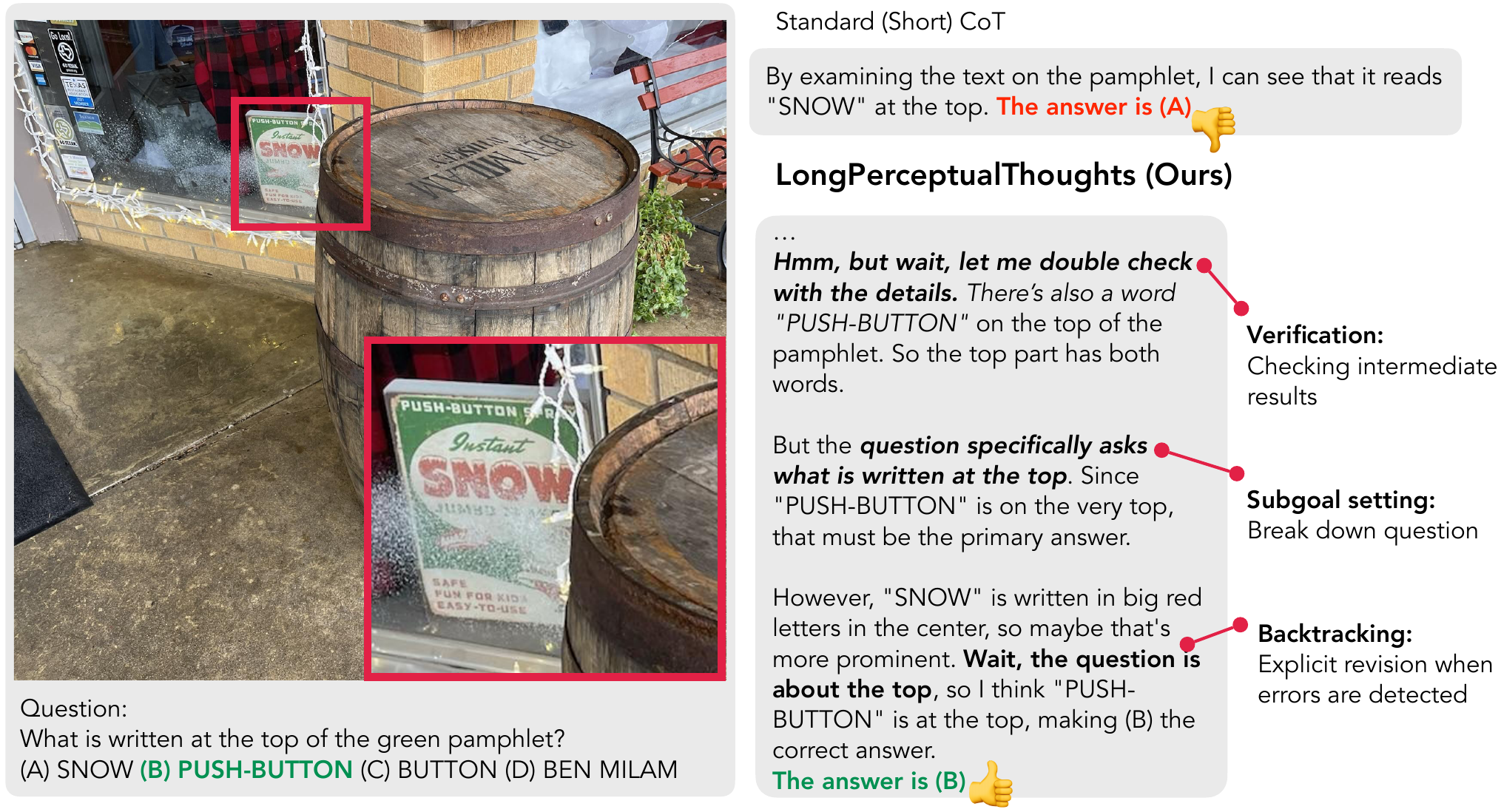}
  \caption{\textbf{\longcot is a new synthetic dataset with 30K long-thought traces for vision-centric tasks.} Each trace contains diverse cognitive behaviors (e.g., verification, subgoal setting, and backtracking), akin to system-2 reasoning.
  % , resulting in long CoT data for vision-centric tasks. 
  CoTs generated by open-source VLMs often produce linear, rigid reasoning traces (top). In contrast, our novel data synthesis framework effectively expands these simple thoughts using frontier reasoning models, equipping VLMs with complex reasoning structures and rich cognitive behaviors—effectively distilling system-2 reasoning into instruction-tuned VLMs.}
\vspace{-0.5cm}

  \label{fig:teaser} 
\end{figure}

\iffalse

\DA{TODO: a bit more meat here, it is in the title now , so we need a bit more of definition here}
\AL{I think it's good to mention V* or any verifier-based paper cuz what we want to do is to includes these reflection/verification/correction into the model.} \DA{yes, please add it, and i will take a pass later} \AL{Added and need more revision. I want to connect the cognitive behaviors in the previous paragraph to vision-centric reasoning. The difference between our work and v${^*}$-like works is that we want the model to reason not by seeing the image better (\ie cropping images), but by thinking in the textual space to find inconsistency, revisiting different image regions, \etc } \AL{Somewhere in the introduce, we need to clarify vision-centric reasoning and visual reasoning.}
\fi

In this work, we take a first step toward a scalable method that synthesizes long CoT data for vision-centric tasks. Specifically, we propose a novel three-stage data synthesis framework that: (1) generates \textit{synthetic} \textit{verifiable} multiple-choice questions from dense image captions, (2) extracts simple CoTs from VLMs for those questions,
and (3) expands these simple CoTs into richer, long-form reasoning traces using frontier reasoning models. Notably, our framework performs \textit{three layers of synthesis:} one to generate questions , one to think, and the last one to think harder.
%\AL{I like saying that we are doing two layers of synthesis. but I think framing the first one as "ask" might be misleading. "one to ask, one to think" sounds like we have two agents talking to each other.}\DA{check}\AL{I got your points now and I think it might be confusing given that we have three steps. It's unclear hich step asks and which step think.}\DA{check}
As shown in Fig.~\ref{fig:teaser}, using our framework, we generate \textbf{\longcot}, a dataset of $30$k examples for both supervised fine-tuning (SFT) and direct preference optimization (DPO), and use it to fine-tune a strong instruction-tuned VLM. The resulting model shows an average +3.4 points improvement across 5 vision-centric benchmarks, including a gain of +11.8 points on V$^*$ Bench, while typical multimodal reasoning datasets fail to improve the base VLM due to overthinking.
Notably, despite being tuned for vision tasks, it also improves on the challenging text reasoning benchmark MMLU-Pro by +2 points.
\section{ Synthesize Long CoT Data for Vision-Centric Tasks}
In this section, we introduce a novel data synthesis framework to synthesize long chain-of-thought (CoT) data for fine-tuning a vision-language model (VLM). Inspired by DeepSeek's R1, we are interested in collecting data consisting of thoughts and answers in the format of \texttt{<think> thought </think> <answer> answer </answer>}. 
We start by discussing two desired properties of reasoning data for vision-centric tasks in Sec.~\ref{sec:long_cot}. 
Based on these two properties, we explain our data synthesis framework in Sec.~\ref{sec:data_pipeline}.
Finally, we use the synthesized long CoT data to construct \longcot, consisting of both SFT and preference datasets in Sec.~\ref{sec:constructing_dataset}.

\subsection{Desired Properties in Long Chain-of-Thought}~\label{sec:long_cot}
%
\iffalse
\DA{move to related maybe?}
While long chains of thought (CoT) typically capture more complex reasoning, vision-language models (VLMs) rarely exhibit such long CoTs in vision-centric reasoning tasks. Directly fine-tuning VLMs on long CoTs may risk disrupting existing capabilities and lead to degraded performance. Prior work in large language models (LLMs) has shown that fine-tuning on familiar data—such as data generated from the VLM with similar size~\citep{li2025smallmodelsstrugglelearn}, from the same model series~\citep{ren-etal-2024-learn}, or data with lower perplexity~\citep{wu2025clearmindsthinkalike}—is crucial for reasoning tasks. We hypothesize that similarly, synthesizing long CoT data with high \textbf{familiarity} to the target VLM is key to effective fine-tuning.
\fi
% 
%While long CoT typically capture more complex reasoning, VLM in vision-centric reasoning problem rarely exhibits long CoT. Directly fine-tunined the model in long CoT may risks destructing the built capabilities leading to inferior performances. Prior works in LLMs have demonstrated that it is essential to fine-tune LLMs on the data that they are familiar with, particularly in reasoning tasks, \eg data generated with similar size~\citep{li2025smallmodelsstrugglelearn}, within the same model series~\citep{ren-etal-2024-learn}, or with a lower perplexity~\citep{wu2025clearmindsthinkalike} \wrt the LLMs beging fine-tuned. We hypothesize that it is also important to synthesize long CoT data with high \textbf{familiarity} with the VLM being fine-tuned. 
% 
% Next, 
Inspired by the recent success in OpenAI's o1 and DeepSeeks's R1, we further define \textit{Long CoT} as an extended, structured rationale that mirrors how a human might approach complex visual reasoning tasks. Unlike the short, linear responses typically produced by current open-source VLMs, Long CoTs  explore alternative solutions, verifying intermediate steps, and adjusting course when necessary.
Drawing on the framework proposed in~\cite{gandhi2025cognitivebehaviorsenableselfimproving}, we characterize Long CoTs in vision-centric tasks through three core cognitive behaviors: \textbf{verification} (checking intermediate conclusions for correctness), \textbf{backtracking} (recognizing and revising failed solution paths), and \textbf{subgoal setting} (breaking down the task into smaller, solvable components). 
% These cognitive behaviors can increase data diversity by providing different ways to approach the same problems.
These cognitive behaviors have been observed in LLM to increase performance by scaling test-time compute~\citep{muennighoff2025s1simpletesttimescaling}.

% Before introducing our data generation pipeline, 
To study cognitive behaviors in vision-centric tasks, we begin by analyzing the outputs of strong instruction-tuned VLMs, following~\cite{gandhi2025cognitivebehaviorsenableselfimproving}.
%, such as Qwen2.5-VL-7B-Instruct. 
Despite its general capabilities, the model rarely displays the cognitive behaviors described earlier.
The responses are often shallow and rigid, lacking the iterative, self-corrective reasoning we aim to capture.
Figure~\ref{fig:behavior_analysis} quantifies this gap between the response from Qwen2.5-VL-7B-Instruct and Gemini 2.0 Flash Thinking. At the end of this section, we introduce \longcot that drastically diversifies the standard CoT with the desired cognitive behaviors.
% \AL{Would it be better to sample the responses frmo Qwen2.5-VL and gemini-flash-thinking? We show the discrepancies of them. And shows that ours bette aligned with gemini-flash stats?}
%Prior work~\citep{gandhi2025cognitivebehaviorsenableselfimproving} validates that such cognitive behaviors are the keys to unlock a self-improving reasoner. To enable strong visual reasoning, we hypothesize that such \textbf{cognitive behaviors} are also important in visual reasoning as well.

\begin{figure}[t]
  \centering
  % Placeholder box (width=6cm, height=4cm)
  %\fbox{\rule{0pt}{8cm} \rule{\textwidth}{0pt}} 
  \includegraphics[width=\textwidth]{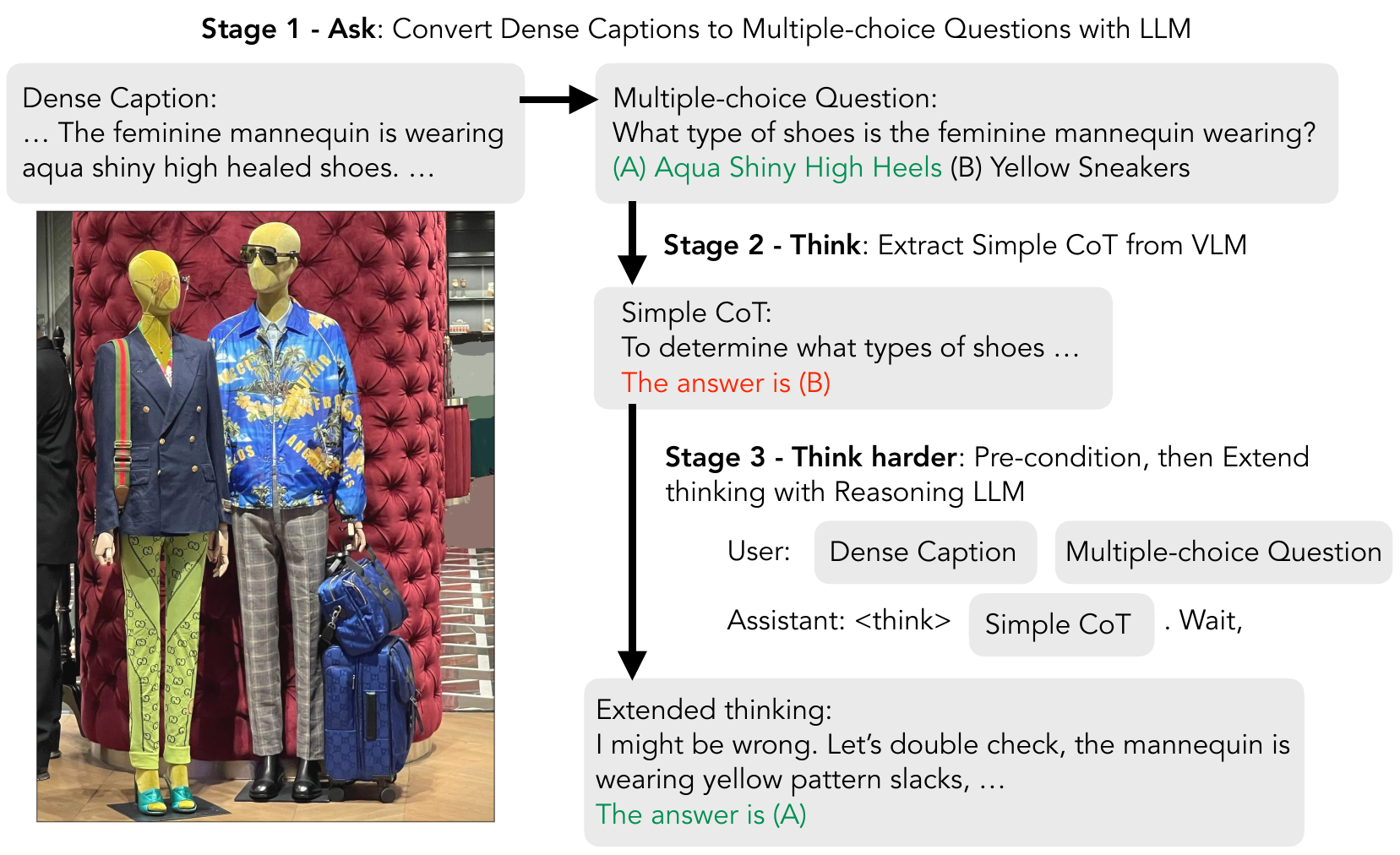}
  \caption{%\textbf{Ask to Generate, Think to Distill, Think Harder to Expand: The three stages to synthesize long CoT data for vision-centric tasks.} 
  \textbf{Ask, Think, and Think Harder: The three stages to synthesize long CoT data for vision-centric tasks.} 
  Assuming the access to an image and its associated dense caption, we first ask an LLM to convert dense captions to multiple-choice questions. In Stage 2, we extract simple CoT from VLM. These simple CoTs typically exhibits shallow and rigid reasoning, especially in vision-centric tasks. Therefore, in Stage 3, we precondition a reasoning LLM with these simple CoTs and append a subtle cue, \eg, ``Wait,'', to elicit more diverse long CoTs.}
  \label{fig:pipeline} 
  %\vspace{-5mm}
  \vspace{-12pt}
\end{figure}

\subsection{Preliminaries}
Formally, given an \textbf{image} $\image$, our goal is to construct a quadruple $(\image, \question, \longthought, \answer)$, consisting of a \textbf{question} $\question$, a long CoT \textbf{reasoning trace} $\longthought$, and a \textbf{final answer} $\answer$. We also assume the access to dense image \textbf{descriptions} $\densecaption$.
A long CoT is composed of multiple 
% reasoning steps 
thoughts
that incorporate cognitive behaviors such as backtracking, verification, and subgoal setting. 
Formally, we define a long CoT as a sequence of intermediate thoughts: $\longthought := \stepthought_1 \oplus \stepthought_2 \oplus \ldots $, where $\oplus$ denotes concatenation and $\stepthought$ is a sequence of sentences, typically delineated by double new lines, \ie ``\textbackslash n\textbackslash n''.
%typically delineated by the period character or double new lines (``.'' or ``\textbackslash n\textbackslash n'').

For preference data—used in reinforcement learning—our goal is to construct a preference pair of $(\image, \question, \longthought^+, \answer^+) \succ (\image, \question, \longthought^-, \answer^-)$
%we extend this to a sixtuple $(\image, \question, \newreasoning^+, \newreasoning^-, \answer^+, \answer^-)$
, where the superscripts $+$ and $-$ indicate the preferred (correct) and non-preferred (incorrect or suboptimal) reasoning trajectories and their answers, and $\succ$ denotes that the left-hand tuple is preferred over the right-hand one.
%, where the superscripts $+$ and $-$ denote the preferred (correct) and non-preferred (incorrect or suboptimal) reasoning trajectories and their corresponding answers, respectively, and $\succ$ denotes that the left-hand tuple is preferred over the right-hand tuple.
%\DA{ and $\succ$ denotes .... }

\iffalse
\begin{equation}
    \newreasoning := \{ \step_1, \step_2 \ldots \}
\end{equation}

A long CoT $\newreasoning$ can be further decomposed as:
\begin{equation}
    \newreasoning := \{ \reasoning, \marker 1, \thought 1,  \marker 2, \thought 2, \ldots \}
\end{equation}
where $\reasoning$ denotes a baseline rationale typically produced by non-commercial VLMs, $m_i$ represents a linguistic marker (e.g., “wait”, “hmm”, etc.) that signals the initiation of a new thought $\tau_i$. Each thought $\tau_i$ is itself a sequence of sentences, typically delineated by the period character (".").
\fi

\subsection{Thought-Expansion: Distilling System-2 Reasoning into Instruction-Tuned VLMs}\label{sec:data_pipeline}

% \textbf{Setup.} 
% Our goal is to collect long CoT to finetune a VLM to reason over the visual details in these images. 

For an image $\image$, we begin by assuming access to its dense image description $\densecaption$ that provides comprehensive  visual features in the image. While in principle, one could also obtain such descriptions using a captioning model, here we assume access to such a dataset \eg, DOCCI~\citep{OnoeDocci2024} or DCI~\citep{Urbanek2023API}. 
In our proposed data synthesis framework, we leverage three foundation models: an LLM , a VLM that takes interleaved image and text as input and generates text, and a reasoning LLM that explicitly produces thoughts and answers. We use $\LLM$, $\VLM$, and $\RLLM$ to denote them, respectively.

\iffalse
We define the thought-expansion process as:
\begin{align}
\{ \reasoning, \marker 1, \thought 1,  \marker 2, \thought 2, \ldots \}:= \model(\reasoning(\image,\question),\densecaption)
\end{align}

Here, $\model(T, C)$ denotes a reasoning LLM that takes as input the initial rationale produced by a VLM for question $\question$ about image $\image$, along with the associated dense caption $\densecaption$. We write $\reasoning(\image, \question) := \VLM(\image, \question)$ to explicitly indicate that $\reasoning$ corresponds to the base model’s rationale (excluding the final answer).
\fi

Below, we describe the three key stages of our data synthesis process. 

% Our goal is to synthesized long CoT  to finetune a VLM to reason over the visual details in these images. 
% We therefore specifically precondition the generation of the the reasoning model to continue from T. 
% To ensure it continues the generation, we sample randomly from a collection of markers and preappended to the generation. 

% Formally, for an image $\image$ and its associated dense caption $\densecaption$, we want to create a triplet containing $( \question, \newreasoning,\answer)$ consisting of a question, a long CoT, and the answer.  $\newreasoning$ consists of a chain of thoughts that incorporates backtracking,verification and subgoal setting.  
% For preference data (RL) we additionally want to associate image and desne caption to the following tuple $( \question, \newreasoning,\answer)$

% For an image $X_v$ and its associated dense caption $X_c$, we want to collect long CoT data consisting of questions, answers, and thoughts $(X_q, X_a, X_{cot})$ to finetune certain VLM.
% \AL{will need to figure out the notation later. I am mixing them now.} \DA{let's define macros, maybe lets consider a  simpler notation. for instance,  I, T, A}

% \DA{overall explanation of the entire pipeline pointing to the method figure before zooming in}

% The QA format can teach VLM to actively attend to visual details for reasoning\DA{why, might be irrelevant?}. 

\textbf{Stage 1: Convert dense descriptions to multiple-choice questions }
% $(\image, \question, \answer^\star)$ using $\LLM$.} 
We first convert dense descriptions into multiple-choice questions (MCQs) using an LLM. Specifically, we prompt $\LLM$ to generate MCQs based on an image and its associated dense descriptions. 
%Specifically, we prompt a language model $\model^\star$ to generate MCQs based on the image $\image$ and its associated dense caption $\densecaption$. 
This step offers two key advantages that are leveraged in subsequent stages: (1) It ensures that each generated question is answerable using only the dense descriptions, allowing us to synthesize the reasoning process purely from the text modality. (2) The multiple-choice format enables easy identification of prediction correctness, which is essential for constructing positive and negative pairs in our preference dataset.
Formally, this step produces a triplet $(\image, \question, \answer^\star) := \LLM(\image, \densecaption)$. 
We use \texttt{gpt-4o-mini} as $\LLM$ to balance the cost and the quality of MCQs.
%In our experiments, we use \texttt{gpt-4o-mini} as $\LLM$, although other model choices are possible.

% In step 1, we obtain $(X_q, X_a)$ and in the following steps, for each question-answering pair, we sample both positive and negative CoT.

%\AL{From "Distill" simple cots to "Extract" simple cots. By "Distill", it makes an impression that we are using different VLMs here. Revised figure 2 accordingly.}
\textbf{Stage 2: Extract Simple CoTs from VLM }
% $(\stepthought_1, \answer_1)$ using $\VLM$.}
To generate long CoTs that the VLM is familiar with, we use the same VLM that will later be fine-tuned. Specifically, we prompt $\VLM$ with the image and the multiple-choice question generated in Stage 1 to produce a rationale and a final prediction, denoted as $(\stepthought_1, \answer_1) := \VLM(\image, \question)$. 
%To generate long CoTs that the VLM is familiar with, we obtain the reasoning traces $\reasoning$ using the same VLM that will later be fine-tuned. Specifically, we prompt the VLM with the multiple-choice questions generated in Step 1 to produce a rationale and a final prediction, denoted as $(\reasoning, \answer) := \VLM(\image, \question)$.
%
Sampling from the same VLM ensures that the synthesized CoTs remain within the model’s output distribution, which we observed to be a key factor in downstream performance.
By comparing the predicted answer $\answer_1$ with the ground-truth answer $\answer^\star$ from Stage 1, we can further categorize the data into positive $(\stepthought_1^+, \answer_1^+)$ or negative examples $(\stepthought_1^-, \answer_1^-)$. These can then be reused to construct either a SFT or a preference dataset. This process is akin to the rejection sampling in self-training algorithms such as RFT~\citep{yuan2023scaling} and STaR~\citep{zelikman2022star}. We choose Qwen2.5-VL-7B-Instruct as our $\VLM$, as the Qwen2.5 series has demonstrated a non-trivial probability of exhibiting cognitive behaviors~\citep{gandhi2025cognitivebehaviorsenableselfimproving}.
% \DA{explain this allows the VLM to stay in track and we keep close of the  VLMs trained distribution}

% \begin{figure}[t]
%   \centering
%   % \includegraphics[width=\textwidth]{assets/figures/cognitive_behaviors.pdf}
%   \includegraphics[width=\textwidth]{assets/figures/dummy_fig2.png}
%   \caption{\textbf{} \AL{gemini numbers are hypothetical.} \AL{This is a figure plot with 3 subplots horiziontally aligned. Each subplot is showing the counts of certain cognitive behaviours. From left to right, it's verification, backtracking, and subgoal setting. In each subplot, we visualize the CoT generated from Qwen2.5-VL with different temperature and our augmented long CoT.}\DA{update dummy figure}} 
%   \label{fig:behavior_analysis}
% \end{figure}

\begin{figure}[t]
    \centering
    %\begin{subfigure}[b]{0.48\textwidth}
    \begin{subfigure}[b]{0.33\textwidth}
        \includegraphics[width=\textwidth]{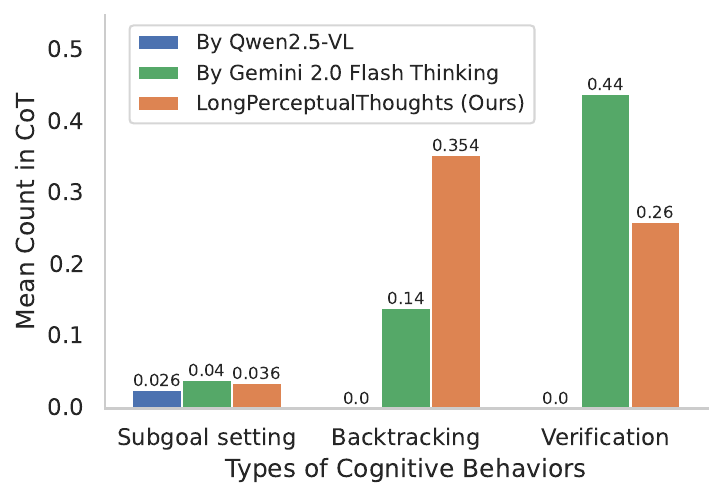}
        \caption{Cognitive behaviors in CoTs}
        \label{fig:behavior_analysis}
    \end{subfigure}
    \hfill
    \begin{subfigure}[b]{0.32\textwidth}
        \includegraphics[width=\textwidth]{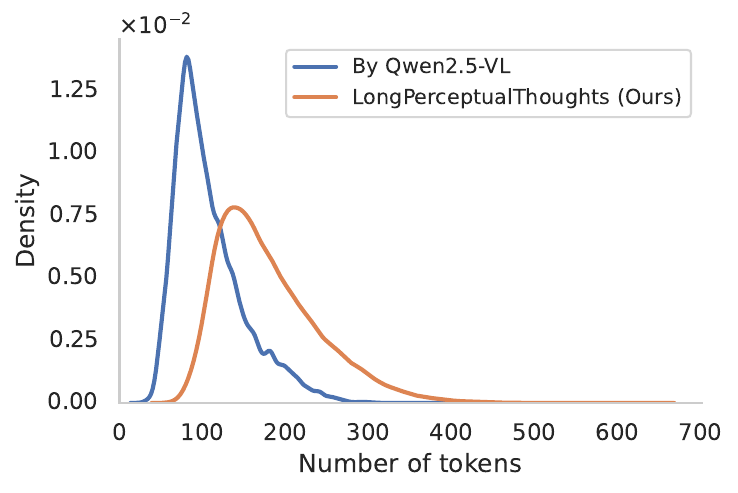}
        \caption{Length of CoTs}
        \label{fig:seq_len_change}
    \end{subfigure}
    \hfill
    %\begin{subfigure}[b]{0.48\textwidth}
    \begin{subfigure}[b]{0.33\textwidth}
        \includegraphics[width=\textwidth]{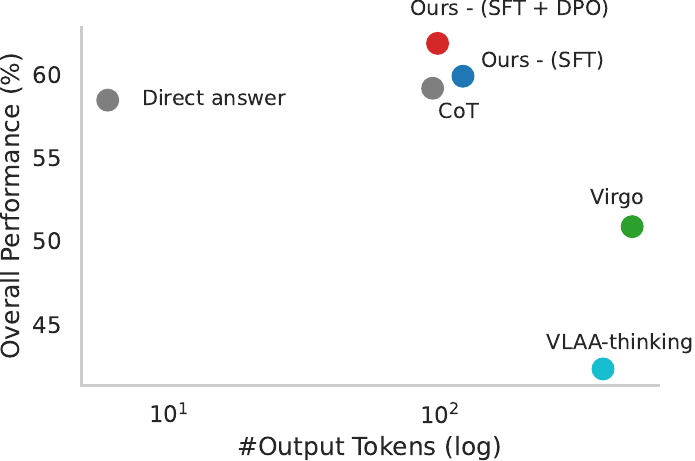}
        \caption{SFT/DPO performances}
        \label{fig:response_length_vs_performance}
    \end{subfigure}
    \caption{
    \textbf{(a) Analysis of Cognitive Behaviors in Chain-of-Thought (CoT).} 
    CoTs from open-source VLMs often follow rigid structures. In contrast, frontier reasoning VLMs—such as Gemini 2.0 Flash Thinking—exhibit more diverse cognitive behaviors, including subgoal setting, backtracking, and verification. Our introduced long CoT dataset, \longcot, also demonstrates a wide range of such behaviors. 
    \textbf{(b) Length of CoTs.} The CoTs in \longcot are significantly longer than those generated by popular VLMs, \eg Qwen2.5-VL.
    %\AL{This might be misleading, we are here comparing in terms of different "dataset generation pipeline", instead of model behaviors.}
    \textbf{(c) Response length vs. aggregated performances.} 
    Fine-tuning VLM on \longcot with complex reasoning structures lead to higher overall performances with slightly more output tokens. On the other hand, fine-tuning on other multimodal reasoning leads to over-thinking and worse performance. Cognitive behaviors are quantified following~\cite{gandhi2025cognitivebehaviorsenableselfimproving}.
    %\DA{right plot, consider just writing Ours(SFT +DPO), Ours (SFT), the entire name + training method makes a bit too long?}
    }
    %\DA{why is the bar of our smaller in suboal setting if it is the same number 0.04?}} \AL{fixed}
    %\label{fig:both}
\end{figure}

% \textbf{Stage 3: Extend thinking $(\stepthought_2, \answer_2)$ using $\RLLM$.}
\textbf{Stage 3: Thought-Expansion using the Reasoning Model.}
The analysis in Fig.~\ref{fig:behavior_analysis} reveals that CoTs sampled from open-source VLMs typically exhibit shallow and rigid reasoning, with limited exploration in the output space. Inspired by the diverse cognitive behaviors observed in the responses of frontier reasoning models, we aim to leverage a reasoning LLM to generate long CoTs.
However, naively sampling from $\RLLM$ can produce CoTs that deviate significantly from the output distribution in VLM, which may degrade the performance of instruction-tuned models during fine-tuning. The similar findings have been discovered in LLMs as well~\citep{ren-etal-2024-learn,li2025smallmodelsstrugglelearn,wu2025clearmindsthinkalike}.
To address this, we introduce a \textit{thought-expansion} mechanism that guides the reasoning LLM $\RLLM$ to extend the CoT produced in Stage 2, while injecting cognitive behaviors such as backtracking, verification, and subgoal setting. Specifically, we precondition $\RLLM$ with the CoT generated by VLM $\stepthought_1$ and append a subtle cue—selected from a set of predefined markers $\marker$ (\eg, “Wait,” “Hmm,” “Alternatively,”)—to elicit more reflective or exploratory responses.
Formally, we structure the prompt as:
% \texttt{User:$\$\densecaption\$ \$\question\$$<STOP>Assistant:<think> $\$\reasoning\$$ Wait,}
%
%
\begin{align*}
%\text{User: } & \texttt{\$}\densecaption\texttt{\$ \, \$}\question\texttt{\$ <STOP>} \\
\text{User: } & \densecaption \oplus \question \\
\text{Assistant: } & \texttt{<think>} \oplus \stepthought_1 \oplus \marker
\end{align*}
and ask $\RLLM$ to continue the thought to obtain $(\stepthought_2, \answer_2)$. This approach enables the reasoning LLM to expand the familiar reasoning traces while enriching them with non-linear problem-solving behaviors. Similar to Stage 2, we can also use $a_2$ to categorize the data into positive or negative examples. Fig.~\ref{fig:pipeline} demonstrates the way to construct such prompt visually. We use DeepSeek-R1-Distill-Qwen-32B as our $\RLLM$, as it is derived from the same Qwen2.5 series as the $\VLM$.
For more details, see the prompt template in Appendix~\ref{appendix:sec:prompts}.

Our proposed framework is scalable and only assumes the access to dense image description datasets. From Stage 1 to Stage 3, for an image $\image$ and its associated descriptions $\densecaption$, we obtain MCQs $(\question, \answer^\star)$ in Stage 1. Then, in Stage 2 and 3, we obtain two intermediate thoughts  and their associated predicted answers $(\stepthought_1, \stepthought_2, \answer_1, \answer_2)$. Finally, we have long CoT data $Z$ obtained by: $\stepthought_1 \oplus \marker \oplus \stepthought_2$. 
We will omit $\marker$ in the following sections for the sake of brevity and clarity. 

% As analyzed in Sec.~\ref{fig:behavior_analysis}, CoT sampled from open-source VLM usually exhibits shallow and rigid reasoning, lacking exploration in the output space. Inspired by the diverse cognitive behaviors observed in the responses of LRM, we want to leverages text-based reasoning model to generate extended CoT $\newreasoning$. However, naively sampling from the LLM might lead to long CoT that is far from the base VLM's  distribution, leading to degraded performance to a well-instruction-tuned VLM. 
% To address the above issue, we introduce a thought-expansion mechanism to guide the LRM to expand the CoT generated by VLM $\reasoning$ with diverse its diverse reasoning capabilities. In particular, we precondition the LLM's output with $\reasoning$ and to encourage cognitive behaviors, we sample from a set of predefined markers, such as ``wait,'' or ``hmm,'',  and then ask LLM to complete the responses. In other words, we precondition the LLM with \texttt{User:$X_c X_q$<STOP>Assistant:<think> $\reasoning$ Wait,} and ask LLM to expand the provided CoT. 

% In this way, for each base CoT $\reasoning$, we effectively expand it to $(\reasoning  \oplus Y_{ext}, Y_a)$\footnote{We use $\oplus$ to denote concatenation.}. Similarly, by comparing the predictions $Y_a$ and $X_a$, we can furhter categorize the data into positive and negative data, $(Y_{cot}^{\{+, -\}} \oplus Y_{ext}^{\{+, -\}}, Y_a)$.
%See more details and the prompt in appendix~\ref{}.

% \DA{explain this is a cost-effective way to inject such behaviours but leveraging already existing models}

% \DA{finished here}

\subsection{Construct SFT and DPO Datasets}\label{sec:constructing_dataset}
% \AL{If necessary, we can expend the motivation of having preference data. See Sec.3.2 in this work~\citep{zhang2025backtracking}}
In Sec.~\ref{sec:data_pipeline}, we described a process to obtain long-form CoTs that not only aligns with the VLM’s output distribution but also contains system-2 reasoning behaviors. 
To construct a supervised fine-tuning (SFT) dataset, we collect CoTs that lead to correct predictions. This includes examples of the form:
\begin{align*}
%(\reasoning^+, \answer), \quad (\reasoning^+ \oplus \thoughts^+, \answer), \quad (\reasoning^- \oplus \thoughts^+, \answer)
(\stepthought_1^+, \answer_1^+), \ (\stepthought_1^+ \oplus \stepthought_2^+, \answer_2^+), \ (\stepthought_1^- \oplus \stepthought_2^+, \answer_2^+)
\end{align*}

%where $\thoughts :=\{\thought 1,  \marker 2, \thought 2, \ldots\}$ denotes the output of the thought-expansion module conditioned on a correct or negative answer (+,-) , and $\oplus$ denotes CoT concatenation.

To construct a preference dataset, we follow \cite{setlur2024rlincorrectsyntheticdata,zhang2025backtracking} and define pairwise preferences based on correctness and compactness~\cite{kimiteam2025kimik15scalingreinforcement}. Specifically:

\textbf{Correctness:}
\begin{align*}
(\stepthought_1^+, \answer_1^+) &\succ (\stepthought_1^-, \answer_1^-) \\
(\stepthought_1^- \oplus \stepthought_2^+, \answer_2^+) &\succ (\stepthought_1^-, \answer_1^-)
\end{align*}

\textbf{Compactness:}
\begin{align*}
(\stepthought_1^+, \answer_1^+) \succ (\stepthought_1^+ \oplus \stepthought_2^+, \answer_2^+)
\end{align*}

Akin to~\cite{setlur2024rlincorrectsyntheticdata}, by constructing the preference pairs of $(\stepthought_1^- \oplus \stepthought_2^+, \answer_2^+) \succ (\stepthought_1^-, \answer_1^-)$, we encourage the model to increase the likelihood $P(\stepthought_2^+, \answer_2^+ | \stepthought_1^-)$  and decrease the likelihood $P(\answer_1^- | \stepthought_1^-)$, leading to better credit assignment.
%\AL{For compactness, we can cite kimi k1.5~\cite{kimiteam2025kimik15scalingreinforcement}}
%\DA{not sure if we need this extra notation, cite them and explain intuitively in a single sentence?}

%\AL{we might want to say that with this two stage data generation approaches, we can perform rough credit assignments. By constructiong the preference pair of $(T^- \oplus T^+, a) \succ (T^-, a)$, we naturally put the blame on the first part of the thought.} \DA{sure, please add it}

\textbf{Filtering.} Since $\stepthought_2$ is generated by a reasoning LLM using dense captions as input, it may include phrases like ``As the description says.'' To address this, we define a list of ``bad words'' and filter out any thoughts containing them. %We also perform deduplication at the end.

\textbf{Details of \longcot.} We use 500 images and their dense captions from DOCCI. Stage 1 produces 4590 multiple-choice questions (MCQs). For long CoT data, we construct an SFT dataset with 30295 examples and a preference dataset with 17208 pairs, following filtering and deduplication. We use \texttt{gpt-4o-mini} as $\LLM$, Qwen2.5-VL-7B-Instruct as $\VLM$, and R1-Distill-Qwen-32B as $\RLLM$.

\section{Experiments}\label{sec:experiments}

In this section, we first describe the experimental setup on five vision-centric benchmarks in Sec.~\ref{sec:setup} and present our main results in Sec.~\ref{sec:main_results}. In Sec.~\ref{sec:beyond_vision}, we go beyond vision by evaluating our fine-tuned VLMs on a challenging text-only benchmark. Lastly, in Sec.~\ref{sec:analysis}, we analyze the response of the fine-tuned VLMs.

% In this section, 
\iffalse
We first describe the experimental setup in Sec.~\ref{sec:setup}. We, then, present the main results in Sec.~\ref{sec:main_results} showcasing the effectiveness of \longcot by comparing it with other visual reasoning datasets. In Sec.~\ref{sec:beyond_vision}, we evaluate fine-tuned VLM on OOD tasks. Lastly, in Sec.~\ref{sec:analysis}, we provide in-depth analysis the VLM fine-tuned on \longcot.
\fi
\subsection{Setup}~\label{sec:setup}
\textbf{Model.} We use Qwen2.5-VL-7B-Instruct~\citep{bai2025qwen25vltechnicalreport} as our base model to fine-tune throughout the paper. For the sake of brevity, we refer to it as \basemodel in this section. We adopt full-parameter fine-tuning using LLaMA-factory~\citep{zheng-etal-2024-llamafactory}. See more training details in Appendix~\ref{appendix:sec:implementation_details}.

\textbf{Benchmarks.}
We evaluate our models on vision-centric tasks. For benchmarks covering general knowledge, we only keep their vision-centric splits, such as MME-RealWorld~\citep{zhang2024mme_real_world} and MMStar~\citep{OnoeDocci2024}. To better clarify the differences, we refer to them as MME-RW-V and MMStar-V, respectively. Additionally, following~\cite{tong2024cambrian}, we include the vision-centric benchmarks: CV-bench, V$^{*}$ Bench, and MMVP, that involve 2D/3D spatial reasoning, fine-trained attribution, coarse scene understanding, \etc 
In total, the benchmarks consist of 10284 images and 15315 questions. More details are in Appendix~\ref{appendix:sec:benchmark_details}.

\textbf{Evaluation metrics.} All the benchmarks used in this work are in multiple-choice question format. We standardize their format and use regex to extract the answers. We report accuracy across all benchmarks.

\textbf{Baselines.} 
To explore the vision-centric capabilities of \basemodel, we evaluate its zero-shot predictions and apply a prompt-based chain-of-thought approach. Specifically, we prompt the model to generate \texttt{<think> thought </think>} before producing an answer—a method we refer to as Internal Thinking CoT.

For multimodal datasets, we compare \longcot with one captioning dataset, DOCCI, and two multimodal reasoning datasets, Virgo~\citep{du2025virgopreliminaryexplorationreproducing} and VAAL-thinking~\citep{chen2025sftrlearlyinvestigation}.
(1) DOCCI is a human-annotated dense caption dataset, highlighting comprehensive descriptions for images. For a fair comparison with \longcot, we use the exact same set of 500 images and their captions as training data. (2) Virgo distills reasoning capabilities from the language model QwQ~\citep{qwq-32b-preview} and the multimodal model QvQ~\citep{qvq-72b-preview}. We adopt Virgo’s self-distillation split, generated by first distilling QwQ into Qwen2-VL-72B-Instruct, then using the fine-tuned model to produce reasoning data for multimodal questions. The Virgo dataset includes $14,540$ examples. (3) VLAA-thinking %, an ongoing project, 
generates multimodal reasoning data by prompting DeepSeek's R1 model with additional caption information. It contains 158k examples, from which we randomly sample 25k for training to match a similar size to our dataset.~\footnote{We accessed the dataset in mid-March 2025.} %\AL{change VL-thinking to VLAA-thinking.}

\begin{table}[t]
\centering
\resizebox{\textwidth}{!}{%
\begin{tabular}{@{}l
>{\columncolor[HTML]{EFEFEF}}r rrrrr@{}}
\toprule
Approach               & \multicolumn{1}{c}{\cellcolor[HTML]{EFEFEF}Avg} & \multicolumn{1}{c}{CV-Bench} & \multicolumn{1}{c}{V$^{*}$ Bench} & \multicolumn{1}{c}{MMVP} & \multicolumn{1}{c}{MMStar-V} & \multicolumn{1}{c}{MME-RW-V}\\ \midrule
 Qwen2.5-VL-7B-Instruct              & 58.47 & 74.74 & 48.51 & 73.67 & 63.73 & 31.68 \\
 + Internal Thinking CoT        & 59.18 & 75.42 & 55.08 & 70.60  & 62.40  & 32.40  \\ 
\midrule \midrule
+ DOCCI                  & 36.14 & 50.82 & 39.96 & 48.67 & 8.67  & 32.58 \\ 
+ VLAA-thinking          & 42.32 & 68.50  & 53.53 & 66.67 & 0.53  & 22.38 \\
+ Virgo                  & 50.87 & 67.22 & 44.14 & 57.67 & 57.60  & 27.71 \\
% + Virgo (our improved version) & 52.58 & 68.94 & 46.54 & 66.33 & 53.47 & 27.60 \\
\midrule \midrule
% DOCCI-MCQ              & 58.92 & \textbf{78.09} & 55.49 & 69.33 & 60.53 & 31.16 \\
% DOCCI-R1-CoT-SFT \textbf{(Ours)}       & 59.9  & 76.05 & \textbf{60.53} & 70    & 60.67 & 32.25 \\
+ \longcot - SFT \textbf{(Ours)}       & 59.90  & 76.05 & \textbf{60.53}  \scriptsize{\textcolor{blue}{(+12.02)}}& 70.00    & 60.67 & 32.25 \\
% DOCCI-R1-CoT-SFT+DPO \textbf{(Ours)}      & \textbf{61.87} & \textbf{76.61} & 60.31 & \textbf{75}    & \textbf{64} & \textbf{33.45} \\ \bottomrule
+ \longcot - SFT + DPO \textbf{(Ours)}      & \textbf{61.87}  \scriptsize{\textcolor{blue}{(+3.4)}} & \textbf{76.61}  \scriptsize{\textcolor{blue}{(+1.8)}}& 60.31 \scriptsize{\textcolor{blue}{(+11.8)}} & \textbf{75.00}  \scriptsize{\textcolor{blue}{(+1.33)}}& \textbf{64.00}  \scriptsize{\textcolor{blue}{(+0.27)}}& \textbf{33.45}  \scriptsize{\textcolor{blue}{(+1.77)}}\\ \bottomrule

\end{tabular}%
}
\caption{\textbf{Main results on five vision-centric benchmarks.} We group the approaches into three categories: training-free methods, existing multimodal reasoning datasets, and our proposed \longcot. On vision-centric tasks, fine-tuning on other multimodal reasoning datasets often leads to poorer performance, likely due to reduced instruction-following ability, domain mismatch, or an inability to capture the complex reasoning learned by larger models. In contrast, fine-tuning on \longcot yields an average improvement of +1.5 points, and this gain increases to +3.4 points when using preference pairs. Notably, it achieves a 12-point improvement on the challenging V$^*$ Bench. 
%\DA{let's put the performance improvement in every benchmark with + in blue and well aligned?}
%\DA{maybe let's do the colour blue in the last one also for v*, it looks a bit weird on the top? and the difference is not to much}
}
\label{tab:main_table}
\end{table}

\subsection{Main Results}\label{sec:main_results}
%\SE{What do you think about splitting this into two parts by treating the OOD text-only results as a separate section?} \AL{In terms of structure, separating them is better. However, given the impressive results that we have in OOD text-only benchmarks, it is worth mentioned earlier. Let me figure out the structure.}\SE{I see, my thought was that it is a somewhat unexpected result so putting it last and independently would make the finding more prominent. I'm not a great paper writer though :D} 
\iffalse
\AL{

Proposed experiment structure:
# Section 3: Experiment
In this section, we first describe the experimental setup on 5 vision-centric benchmarks in Sec 3.1 and describe our main results in Sec 3.2. In Sec. 3.3, we goes beyond vision, and evaluate our finetuned model on challenging text-only benchmark. Lastly, in section 3.4, we analyze the responses from the fine-tuned VLM.
Section 3.1: setup
Model
Benchmarks
Evaluation metrics
Baselines
Section 3.2: main results
think longer, see better
Dense captions, sparse gains
Too much to digest
Section 3.3 Beyond vision: evaluation OOD tasks
What MMLU-Pro is
Results
Section 3.4: Analaysis of Model Responses
---
We want to give a “hint” that we perform not only on vision centric tasks, but also on OOD text only benchmarks.
We want to make text-only part as a standalone section so that we can describe the dataset and the results without mixing it with any vision-centric setup

}
\fi

We report aggregated performances in Table~\ref{tab:main_table} and detail our main findings on five vision-centric benchmarks:

%\textbf{Think Longer, See Better: \longcot consistently and notably improves performance on vision-centric benchmarks} 
\textbf{\longcot consistently improves performance on vision-centric benchmarks by +3.4 points via DPO.}
%\DA{discuss best bragging wao results wrt to base model first, then what didnt work to well and our hypothesis. needs a bit of clarification}
% , especially after preference-based fine-tuning, \eg DPO~\citep{rafailov2023direct}.} 
% Structure
% - Observations 
%   - Other reasoning datasets Virgo and VL-Thinking hurt performance of the base model.
%   - Instead, \longcot improves upon baseline CoT by xxx. Since we are able to construct preference pairs we can boost performance further using DPO.
% - Explanations
We first perform supervised fine-tuning on the synthesized \longcot. Across 5 benchmarks, we improve \basemodel by nearly +1.5 points on average, especially in challenging tasks such as V$^*$ bench, improving by +12 points. However, the improvements on the rest of the benchmarks are marginal. We hypothesize that this is due to noisy or erroneous tokens in our SFT datasets, which may hurt fine-tuning performance.
%We hypothesize that this is because our SFT datasets contains traces containing noisy erroneous tokens and fine-tuning on them might be detrimental
While several prior works suggest the impacts of such erroneous tokens are marginal, they either work on models under 300M parameters~\citep{ye2024physicslanguagemodels22} or target different aspects, such as safety alignment~\citep{zhang2025backtracking}. In this work, we try not to over-engineer the training recipe to highlight the effectiveness of the synthesized datasets.
Unlike VLAA-thinking and Virgo that simply distill knowledge from reasoning LLMs or VLMs, our data generation pipeline allows us to construct preference data. By fine-tuning on these preference pairs, the aforementioned erroneous tokens might naturally be mitigated. For example, by performing preference-based fine-tuning such as DPO, on $(\stepthought_1^- \oplus \stepthought_2^+, \answer_2^+) \succ (\stepthought_1^-, \answer_1^-)$, the model should naturally increase the likelihood of $P(\stepthought_2^+, \answer_2^+ | \stepthought_1^-)$ as opposed to $P(\answer_1^- | \stepthought_1^-)$. This helps the model reduce the impact of erroneous tokens. %Overall, 
We find that by first applying SFT and then DPO, we obtain consistent improvements across all evaluation datasets, improving by +3.4 accuracy points.
%\DA{this previous sentence should be on the top. second sentence, people wont read this far and stop after they read 1.5, which doesnt look as exciting.}
% 
Overall, we find that \longcot generally leads to consistent improvements and the preference data is the key to bring up the improvements.

%\textbf{Thinking Outside the \textit{Camera Frame}: \longcot improves OOD performance on MMMLU-Pro (Text-Only) }: As shown in Fig.~\ref{tab:beyond_vision}, we find that \basemodel fine-tuned on \longcot surprisingly improves on these text-only reasoning tasks, with an average gain of 2 points. In contrast, VL-thinking and Virgo hurt performances, suggesting that directly distilling from stronger teachers may lead to sharp drops in OOD tasks. We propose two hypotheses for \longcot’s effectiveness: (1) it introduces complex reasoning structures that improve \basemodel’s general reasoning abilities; and (2) it remains close to the original output distribution, making the new reasoning skills easier to learn without disrupting existing knowledge. Additional MMLU-Pro evaluation details are provided in Appendix\ref{sec:}.\DA{this is very strong lets put it here, please clarify and improve readability?}

\iffalse
\textbf{Vision-centric tasks benefit from reasoning.} On average, performance of \basemodel slightly improves by 0.7 accuracy points. While it is small and has been shown in prior works~\citep{}, it suggests that vision-centric tasks can improve via stronger reasoning capabilities. This encourages us to further push the perception capabilities through better reasoning. \DA{message needs clarity, consider removing this}
\fi

% \textbf{Off-the-shelf Dense caption dataset hurts instruction-tuned VLMs.} 
%\textbf{Dense Captions, Sparse Gains: Off-the-shelf captioning data hurts instruction-tuned VLMs on vision-centric benchmarks.}
\textbf{Off-the-shelf captioning data hurts instruction-tuned VLMs on vision-centric benchmarks.}
Since \longcot is derived from DOCCI, we are interested to see if fine-tuning \basemodel on DOCCI improves. Table~\ref{tab:main_table} shows that training on DOCCI results in inferior performances. Perhaps expected, we find that fine-tuning on DOCCI alone especially leads to bad instruction following. 
%\SE{I feel this paragraph should definitely be above the MMLU-Pro results as it refers to Table 1 only?} \DA{sure, please move it}

%\textbf{Too Much to Digest: Off-the-shelf distillation hurts performance on vision-centric benchmarks.} 
\textbf{Off-the-shelf distillation hurts performance on vision-centric benchmarks.} Both Virgo and VLAA-thinking are multimodal reasoning datasets. VLAA-thinking is distilled from R1 with the help of image captions. Virgo is distilled from a fine-tuned VLM distilled from QwQ. While both datasets are equipped with complex reasoning structures, fine-tuning the \basemodel does not improve vision-centric performance; instead, finetuning on VLAA-thinking and Virgo hurts the performances by -16.15 and -7.6 points, respectively. We hypothesize that there are two reasons that lead to the performance drops: (1) Both datasets are distilled from a much much larger LLMs (671B R1 model) or VLMs (Qwen2-VL-72B-Instruct), potentially resulting in the learnability gap~\citep{li2025smallmodelsstrugglelearn}. (2) In particular, the multimodal reasoning data from Virgo is math-focused. We hypothesize that there is a gap between reasoning over perceptual tasks and math-related tasks. On the other hand, VLAA-thinking consists of a diverse set of datasets including DocVQA~\citep{docvqa}, ChartVQA~\citep{masry-etal-2022-chartqa}, %ArXivQA~\citep{li-etal-2024-multimodal-arxiv}, 
\etc When using reasoning data exclusively from more natural image sources, %—specifically ALLaVA-LAION~\citep{chen2024allava} and VizWiz~\citep{gurari2020captioning}—
we surprisingly observe worse performance than random subsampling. See Appendix~\ref{appendix:sec:vl_thinking_similar_images} for details.

\subsection{Beyond Vision: Evaluation on the Text-Only Reasoning Benchmark}\label{sec:beyond_vision}
Following the same setup in Sec.~\ref{sec:setup}, we evaluates VLMs fine-tuned on multimodal reasoning training datasets in out-of-distribution (OOD) tasks. In particular, we adopt MMLU-Pro, a challenging text-only reasoning benchmark.
%\DA{We emphasize this benchmark is OOD for the model. \longcot is a vision datasets where all examples contain images.}

\iffalse
    \begin{table}[t]
    \centering
    \input{assets/table-beyond-vision}
    \caption{\textbf{Evaluation on text-only reasoning benchmark: MMLU-Pro.} MMLU-Pro is a challenging text reasoning benchmark. Although \longcot is designed for vision-centric tasks, we find its reasoning capabilities surprisingly transferable to text-only tasks. In contrast, fine-tuning on other multimodal benchmarks significantly degrades performance compared to the base model.}
    % MMLU-Pro is a challenging text reasoning benchmark. While \longcot is designed to improve vision-centric tasks, perhaps surprisingly, we find that the reasoning capabilities captured in \longcot can be further reused in text-only reasoning tasks. In sharp contrast, fine-tuning on other multimodal benchmarks leads to sharp performance drop as compared to the base model.
    \label{tab:beyond_vision}
    \end{table}
\fi

\begin{wraptable}{r}{0.4\textwidth}
    \vspace{-3mm}
    \centering
    \begin{tabular}{@{}lr@{}}
        \toprule
        Approach                    & \multicolumn{1}{c}{Acc} \\ \midrule
        Qwen2.5-VL-7B-Instruct              & -\\
        + CoT                             & 48.07\\ 
        \midrule \midrule
        + DOCCI                          & 32.99\\ 
        + VLAA-thinking                  & 21.56\\
        + Virgo                          & 37.95\\
        \midrule \midrule
        + \textbf{Ours} - SFT          & \textbf{50.77} \\ 
        + \textbf{Ours} - SFT + DPO    & 50.20 \\ 
        \bottomrule
    \end{tabular}
    \caption{\textbf{Evaluation on out-of-distribution tasks text-only reasoning benchmark MMLU-Pro.}} %MMLU-Pro is a challenging text reasoning benchmark. Although \longcot is designed for vision-centric tasks, we find its reasoning capabilities surprisingly transferable to text-only tasks. In contrast, fine-tuning on other multimodal benchmarks significantly degrades performance compared to the base model.}
    \label{tab:beyond_vision}
    \vspace{-3mm}
\end{wraptable}

\textbf{MMLU-Pro~\citep{wang2024mmlupro}.} MMLU-Pro is built on top of MMLU~\citep{hendrycks2021measuring} by integrating more reasoning-focused questions and expanding the choices set. MMLU-Pro spans 14 diverse domains including mathematics, physics, chemistry, \etc, encompassing over 12000 questions. 

\textbf{Results.} 
%\textbf{Thinking Outside the \textit{Camera Frame}: \longcot improves OOD performance on MMMLU-Pro (Text-Only) }: 
As shown in Table.~\ref{tab:beyond_vision}, we find that \basemodel fine-tuned on \longcot surprisingly improves on these text-only reasoning tasks, with an average gain of +2 points. In contrast, VLAA-thinking and Virgo hurt performance, suggesting that directly distilling from stronger teachers may lead to sharp drops in OOD tasks. We propose two hypotheses for LongPerceptualThoughts’ effectiveness: (1) it introduces complex reasoning structures that improve \basemodel’s general reasoning abilities; and (2) it remains close to the original output distribution, making the new reasoning skills easier to learn without disrupting existing knowledge. Additional MMLU-Pro evaluation details are provided in Appendix~\ref{appendix:sec:mmlu_pro_full_results}.

%\DA{this is very strong lets put it here, please clarify and improve readability?}
%\vspace{-3.1mm}
\subsection{Analysis on Fine-tuned VLM Responses}\label{sec:analysis}
To better understand our fine-tuned VLM, we visualize its responses with two key factors: aggregated performances and question difficulties. 

\textbf{Response length vs. performances.}
There has been growing interest in how LLMs leverage test-time compute. To investigate this, we aggregated response lengths and performance across five vision-centric benchmarks.
Fig.~\ref{fig:response_length_vs_performance} illustrates the relationship between test-time compute—measured by response length—and model performance. We observe that VLMs fine-tuned on \longcot tend to produce slightly longer responses, especially after SFT. Interestingly, DPO training results in shorter responses, which aligns with the compactness encouraged during DPO pair construction. One possible direction is to exclude such preference pairs to allow models to make fuller use of test-time compute.
In contrast, other multimodal reasoning benchmarks reveal signs of overthinking, where models generate unnecessarily lengthy responses.
%\AL{might need one closing sentence.}

%There has been tremendous interests in how LLM leverages test-time compute. We aggregate the response length and performances across all five vision-centric benchmarks we studied. Fig.~\ref{fig:response_length_vs_performance} demonstrate the relation between test-time compute, \ie response length, and their performances. We find that VLMs fine-tuned on \longcot do exhibit slightly longer responses, especially after SFT. And, interestingly DPO training leads to short responses, which makese sense since when constructing the DPO pairs, we encourage compactness in the model responses. One could explore to exclude such preference pairs in DPO training to potentially using more test-time compute. On the other hand, other mutlimodal reasoning benchmarks leads to over-thinking behaviors. 

\textbf{Response length vs. question difficulty.} 
Another desirable characteristic of the thinking process in LLMs is their ability to allocate appropriate “thinking time” based on a question’s difficulty.
Following prior works~\citep{lightman2024lets,snell2025scaling}, we define question difficulty with respect to a base VLM, i.e., Qwen2.5-VL-7B-Instruct. For each question, we estimate the model's accuracy using 11 samples and bin the questions into two quantiles: easy and hard. Our analysis focuses on the outputs of the VLM fine-tuned via DPO on \longcot.
We observe that the model naturally allocates more test-time compute—reflected in longer responses—for harder questions, where its original (pre-fine-tuning) performance was worse. See Appendix~\ref{appendix:sec:response_length} for details and visualization.
%Fig~\ref{fig:seq_len_vs_difficulty} shows the distribution of response lengths across the easy and hard bins for four different tasks. We observe that the model naturally allocates more test-time compute—reflected in longer responses—for harder questions, where its original (pre-fine-tuning) performance was worse. 
%\AL{downplay this. Put in appendix.}

% \subsection{Discussion}

\vspace{-2mm}
\section{Related Work}
\vspace{-1mm}
% \subsection{Reasoning in Large Language Models}
\textbf{Reasoning in Large Language Models.}
The complex reasoning abilities of large language models (LLMs) have been uncovered through various approaches. Chain-of-thought (CoT) prompting elicits their intrinsic reasoning capabilities, improving performance on language-based causal reasoning tasks~\citep{cot}, and has been extended into tree-based searches to enhance reasoning further~\citep{tree_of_thought}. Similar search-like behavior can be induced through verifier guidance~\citep{lifshitz2025multiagentverificationscalingtesttime}, curated datasets~\citep{shao2024deepseekmathpushinglimitsmathematical}, or supervised fine-tuning on synthetic reasoning data~\citep{stream_of_search_sos,beyond_a_star}.
More recently, DeepSeek-R1~\citep{guo2025deepseek} achieved state-of-the-art reasoning through reinforcement learning, exhibiting human-like traits such as self-correction and verification. In contrast, s1~\citep{muennighoff2025s1simpletesttimescaling} improves mathematical reasoning via supervised fine-tuning on 1000 distilled reasoning traces from frontier models.
While most prior work focuses on math and coding tasks, our goal is to explore how such strong reasoning capabilities can be effectively applied to perception.
% -centric problems.

%Moreover, SFT using deliberated long reasoning VQA dataset may significant improve the reasoning ability in widespread domains including math, coding, path searching, \etc ~\cite{muennighoff2025s1simpletesttimescaling} provide 1K high-quality reasoning dataset among dozens of knowledge domains and boost LLM's math ability. SoS~\citep{stream_of_search_sos} generates synthetic mathematical calculation game dataset to advance the LLM performance on such games.~\cite{beyond_a_star} discovers the best path searching problem by SFT of synthetic reasoning dataset given the ground truth paths. Apart from using offline static dataset, RL introduces verifiable reward models to teach LLM by online reward backpropagation. Other methods introduce outcome and process reward models or using verifiers are also proven to improve reasoning ability~\citep{liao2024feedbackenhancesemanticgrounding,lifshitz2025multiagentverificationscalingtesttime}.

% \begin{enumerate}
%     \item Prompt to reason: Chain-of-thought~\citep{cot}, tree-of-thought~\citep{tree_of_thought}, 
%     \item SFT to reason: s1~\citep{muennighoff2025s1simpletesttimescaling}, SoS~\citep{stream_of_search_sos}, Beyond A*~\citep{beyond_a_star}, DeepSeek-R1-Distilled.
%     \item RL with verifiable rewards to reason: DeepSeek-R1.
%     \item Others: outcome/process rewards models, verifiers~\citep{liao2024feedbackenhancesemanticgrounding,lifshitz2025multiagentverificationscalingtesttime}
% \end{enumerate}

% \subsection{Reasoning in Vision-Centric Tasks}
\textbf{Reasoning in Vision-Centric Tasks.}
Unlike reasoning in math or coding tasks, vision-centric problems often involve significant uncertainty due to partial information, perceptual noise, and visual ambiguities. Prior works primarily address this by helping VLMs "see" better. For instance, SEAL~\citep{v_star} uses a search-like cropping mechanism to iteratively navigate an image, while VisualCoT~\citep{visual_cot} generates auxiliary visual cues to guide attention. %Visual Sketchpad~\citep{hu2024visual} overlays visual markers to help models visualize the consequences of actions described in text. 
Other approaches~\citep{wang2024exovip,liao2024feedbackenhancesemanticgrounding} decompose complex tasks into simpler verification steps to enhance model robustness.
In contrast, we aim to teach VLMs to reason better—encouraging them to explore multiple solution paths by revisiting image regions, verifying intermediate conclusions, and engaging in textual inner monologue. 
Concurrent work on multimodal reasoning addresses this challenge, particularly in math problem solving, using techniques such as distillation from advanced reasoning LLMs~\citep{du2025virgopreliminaryexplorationreproducing,thawakar2025llamavo1} and reinforcement learning~\citep{liu2025visual,huang2025visionr1incentivizingreasoningcapability}.
In this work, we study how system-2 reasoning can improve vision-centric tasks, and propose a data synthesis framework that generates long CoT examples to teach visual reasoning through deliberate, step-by-step thinking in the textual space.

\section{Conclusions}
In this work, we explore how system-2 reasoning can enhance vision-centric tasks. We introduce a novel, scalable data synthesis framework that requires only dense image captions. The framework generates verifiable multiple-choice questions, extracts simple chains of thought (CoTs) from vision-language models (VLMs), and expands them into rich, long-form reasoning traces using frontier reasoning models. This process yields \longcot, a synthetic dataset containing 30k detailed reasoning traces for perceptual tasks. Fine-tuning Qwen2.5-VL-7B-Instruct on \longcot improves performance by +3.4 points across five vision benchmarks, including ann +11.8-point gain on   V$^{*}$ Bench. Remarkably, despite being trained on vision tasks, the   model also improves  by +2 points on the out-of-distribution text-only reasoning benchmark MMLU-Pro.

\section*{Acknowledgements}

We thank Rafid Mahmood, Jaehun Jung, Jen-Hao Cheng, Ali Hatamizadeh, Ximing Lu, Hyunwoo Kim and Amlan Kar for their helpful comments and feedback on an early discussions and draft of this paper.

\clearpage

\bibliography{reference}
\bibliographystyle{colm2025_conference}

\clearpage

\appendix

\section{Table of Content}

\begin{enumerate}
    \item Sec.~\ref{appendix:sec:benchmark_details} elaborates the details of the considered five vision-centric benchmarks
    \item Sec.~\ref{appendix:sec:additional_results} provides additional experimental results including additional comparison with self-training and our efforts to improve VLAA-thinking and Virgo. 
    \item Sec.~\ref{appendix:sec:mmlu_pro_full_results} provides the full evaluation results on text-only reasoning benchmark, MMLU-Pro.
    \item Sec.~\ref{appendix:sec:implementation_details} provides implementation details in dataset generation, VLM training, and VLM inference.
    \item Sec.~\ref{appendix:sec:qualitative_example} provides additional qualitative results of our dataset generation pipeline.
    \item Sec.~\ref{appendix:sec:response_length} provides the analysis of fine-tuned VLM's response length versus question difficulties.
\end{enumerate}

\iffalse
TODOs
\begin{itemize}
    % \item Include DOCCI-MCQ results??
    % \item MMLU-Pro detailed results: \label{appendix:sec:mmlu_pro_experiments} \SE{On it}
    % \item Additional experiment results of DOCCI-MCQ-CoT (rejection samplgin FT)
    \item A few qualitative examples? (dataset samples)
    \item Fill prompts used.
\end{itemize}
\fi

\section{Benchmark details}\label{appendix:sec:benchmark_details}

\begin{table}[t]
\centering
\resizebox{\textwidth}{!}{%
\begin{tabular}{@{}lrrrrr
>{\columncolor[HTML]{EFEFEF}}r}
\toprule
             & \multicolumn{1}{c}{CV Bench} & \multicolumn{1}{c}{V$^{*}$ Bench} & \multicolumn{1}{c}{MMVP} & \multicolumn{1}{c}{MMStar-V} & \multicolumn{1}{c}{MME-RealWorld-V}   & \multicolumn{1}{c}{\cellcolor[HTML]{EFEFEF}Total}   \\ \midrule
\# Images &                                   2638&     191                            &      300                               &   750                           &               6405        &  10284 \\ 
\# Questions &                                   2638&     191                            &      300                               &   750                           &               11436        &  15315 \\ \bottomrule
\end{tabular}%
}
\caption{\textbf{Vision-centric benchmark statistics.}}
\label{appendix:tab:benchmark_stats}
\end{table}

We describe the details of each benchmark:

\begin{enumerate}
    \item CV-Bench~\citep{tong2024cambrian} is a comprehensive benchmark of over 2k manually-inspected examples, evaluating visual understanding across domains such as object recognition, scene understanding, and visual reasoning.
    \item V$^{*}$ Bench~\citep{v_star} targets fine-grained visual reasoning tasks that demand detailed analysis of visual elements.
    \item MMVP~\citep{mmvp} tests visual pattern recognition using “CLIP-blind pairs”—visually distinct images perceived as similar by CLIP—highlighting systematic limitations in VLMs.
    \item MMStar-V includes tasks assessing instance-level reasoning, fine-grained perception (detecting subtle visual details), and coarse perception (understanding overall scene context).
    \item MME-RW-V. MME-RealWorld evaluates real-world visual understanding across domains such as autonomous driving, remote sensing, monitoring, diagrams, tables, and OCR, testing both perception and reasoning. From these, we select three perception tasks—Remote Sensing, Monitoring, and Autonomous Driving—and two reasoning tasks—Monitoring and Autonomous Driving—to form MME-RealWorld-V.
\end{enumerate}

As a result, our evaluation provides a comprehensive view on the perceptual capabilities enabled by the training datasets under consideration. Table~\ref{appendix:tab:benchmark_stats} shows the basic statistics of the considered benchmarks.

\section{Response length vs. question difficulty}\label{appendix:sec:response_length}

\begin{figure}[t]
    \centering
    \begin{subfigure}[b]{0.32\textwidth}
        \includegraphics[width=\textwidth]{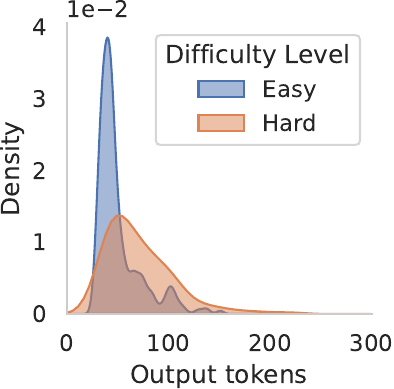}
        \caption{\scriptsize{CV-Bench - Count}}
    \end{subfigure}
    % \hfill
    % \begin{subfigure}[b]{0.24\textwidth}
    %     \includegraphics[width=\textwidth]{assets/figures/sequence_length_vs_difficulty/full_mme_real_world_variants_answer_perception_monitoring_2196.pdf}
    %     \caption{\scriptsize{MME-RW - Monitoring}}
    % \end{subfigure}
    \hfill
    \begin{subfigure}[b]{0.32\textwidth}
        \includegraphics[width=\textwidth]{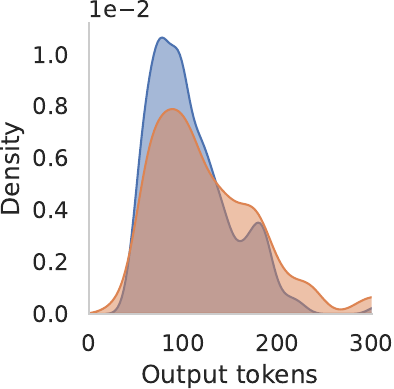}
        \caption{\scriptsize{MMStar - InstReason}}
    \end{subfigure}
    \hfill
    \begin{subfigure}[b]{0.32\textwidth}
        \includegraphics[width=\textwidth]{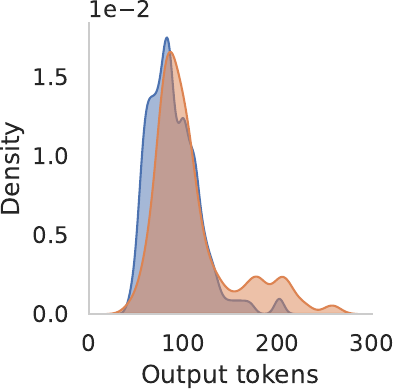}
        \caption{\scriptsize{MMStar - CoarsePercep}}
    \end{subfigure}
    \caption{\textbf{Response lengths vs. question difficulties.}
    We analyze the responses of the VLM fine-tuned on \longcot via DPO. Interestingly, we find that the model finetuned in our data naturally allocates more test-time compute for hard questions. We follow~\cite{lightman2024lets,snell2025scaling} and determine question complexity using rollouts on the base model.
    %\AL{To replace the figure. Add references for difficulty measurements.}
    %We follow~\cite{lightman2024lets,snell2025scaling} and determine question complexity using rollouts on the base model.
    % where the VLM performs worse before fine-tuning.
    %Distribution of response lengths in terms of token counts for easy and hard questions for different data subsets.
    %\DA{awesome, let's make it selfcontained}
    }
    \label{fig:seq_len_vs_difficulty}
\end{figure}

Following prior works~\cite{}, we define question difficulty with respect to a base VLM, \ie Qwen2.5-VL-7B-Instruct. For each question, we estimate the model's accuracy using 11 samples and bin the questions into two quantiles: easy and hard. Our analysis focuses on the outputs of the VLM fine-tuned via DPO on \longcot.
Fig~\ref{fig:seq_len_vs_difficulty} shows the distribution of response lengths across the easy and hard bins for four different tasks. We observe that the model naturally allocates more test-time compute—reflected in longer responses—for harder questions, where its original (pre-fine-tuning) performance was worse.

\section{Additional Results}\label{appendix:sec:additional_results}

%\paragraph{Compared with Rejection Sampling Fine-Tuning~\citep{yuan2023scaling}}\label{appendix:sec:vl_thinking} \AL{To fill the details}

\begin{table}[t]
\centering
\resizebox{\textwidth}{!}{%
\begin{tabular}{@{}l
>{\columncolor[HTML]{EFEFEF}}r rrrrr@{}}
\toprule
Approach               & \multicolumn{1}{c}{\cellcolor[HTML]{EFEFEF}Avg} & \multicolumn{1}{c}{CV-Bench} & \multicolumn{1}{c}{V$^{*}$ Bench} & \multicolumn{1}{c}{MMVP} & \multicolumn{1}{c}{MMStar-V} & \multicolumn{1}{c}{MME-RW-V}\\ \midrule
Qwen2.5-VL-7B-Instruct              & 58.47 & 74.74 & 48.51 & 73.67 & 63.73 & 31.68 \\
 % + Internal Thinking CoT        & 59.18 & 75.42 & 55.08 & 70.6  & 62.4  & 32.4  \\ 
\midrule \midrule
% + DOCCI                  & 36.14 & 50.82 & 39.96 & 48.67 & 8.67  & 32.58 \\ 
VLAA-thinking                     & 42.32 & 68.50 & 53.53 & 66.67 & 0.53  & 22.38 \\
+ only natural images           & 34.96 & 61.91 & 28.86 & 55.00 & 6.20  & 22.86 \\
\midrule \midrule
Virgo                           & 50.87 & 67.22 & 44.14 & 57.67 & 57.60  & 27.71 \\
+ improved formatting     & 52.58 & 68.94 & 46.54 & 66.33 & 53.47 & 27.60 \\
% DOCCI-MCQ              & 58.92 & \textbf{78.09} & 55.49 & 69.33 & 60.53 & 31.16 \\
% DOCCI-R1-CoT-SFT \textbf{(Ours)}       & 59.9  & 76.05 & \textbf{60.53} & 70    & 60.67 & 32.25 \\
% + \longcot - SFT \textbf{(Ours)}       & 59.9  & 76.05 & \textbf{60.53} & 70    & 60.67 & 32.25 \\
% DOCCI-R1-CoT-SFT+DPO \textbf{(Ours)}      & \textbf{61.87} & \textbf{76.61} & 60.31 & \textbf{75}    & \textbf{64} & \textbf{33.45} \\ \bottomrule
% + \longcot - SFT + DPO \textbf{(Ours)}      & \textbf{61.87}  \textcolor{blue}{(+3.4\%)} & \textbf{76.61} & 60.31 & \textbf{75}    & \textbf{64} & \textbf{33.45} \\ 
\bottomrule
\end{tabular}%
}
\caption{Attempted improvements on top of VLAA-Thinking and Virgo baselines. }%\AL{I like the presentation here!}}
\label{tab:attemped_improvements_baselines}
\end{table}
\paragraph{VLAA-Thinking and Virgo adjustments.}
\label{appendix:sec:vl_thinking_similar_images}
As we saw degradation in performance when training on both, Virgo and VLAA-Thinking, we spent additional time investigating the datasets and the model behavior they are causing which lead to these results.

We found that VLAA-Thinking consists of large proportions of math questions whereas natural image data is dominating the considered benchmarks as we focus on perceptual tasks. We hypothesize that this distribution shift might lead to lower performance. To investigate, we consider a version of VLAA-Thinking where we only keep the image subsets containing natural images, \ie, ALLaVA-LAION and VizWiz, and randomly sample a subset of the same size. For Virgo, we found that predictions would not consistently respect the system prompt when formatting answers leading to inconsistencies with our regex-based evaluation. We thus explore a version of the dataset where we only copy the answer provided inside \texttt{\textbackslash{}boxed\{\}} into \texttt{<answer>} tags, discarding the justification part of the answer, while keeping the thinking part of the dataset the same.

The results of both adjustments can be found in Table \ref{tab:attemped_improvements_baselines}. We observe that training on only natural images in VLAA-Thinking hurts performance further, likely due to the limited data diversity. One the other hand, when applying improved answer formatting the results on Virgo improve slightly from $50.87\%$ to $52.58\%$. However, despite these adjustments, the datasets still fail to improve beyond the base model.

\section{Full MMLU-Pro Evaluation Results}~\label{appendix:sec:mmlu_pro_full_results}
We provide the detailed results on all MMLU-Pro categories in Table~\ref{appendix:tbl:mmlu-pro-full}. We observe that the model fine-tuned on our \longcot dataset consistently outperforms the baselines and provides improvements on top of the base model except for the \texttt{Other} category.

\begin{table}
    \centering
    \resizebox{\textwidth}{!}{%
    
\begin{tabular}{@{}l
    >{\columncolor[HTML]{EFEFEF}}rrrrrrrrrrrrrrr}
    \toprule
     & \multicolumn{1}{c}{\cellcolor[HTML]{EFEFEF}Avg} & Biology & Business & Chemistry & CompSci. & Econ. & Engin. & Health & History & Law & Math & Phil. & Physics & Psych. & Other \\
    \midrule 
    Qwen2.5-VL-7B-Instruct & 48.07 & 68.62 & 55.77 & 44.79 & 49.51 & 61.26 & 34.26 & 47.68 & 43.57 & 24.89 & 50.41 & 38.88 & 47.19 & 60.65 & \bfseries 45.56 \\
    \midrule \midrule
    DOCCI & 32.99 & 51.60 & 42.33 & 22.61 & 37.32 & 43.48 & 18.89 & 32.76 & 22.31 & 10.26 & 40.19 & 29.46 & 25.56 & 51.13 & 33.98 \\
    VLAA-Thinking & 21.56 & 25.24 & 27.76 & 15.11 & 20.73 & 25.47 & 7.64 & 24.45 & 29.40 & 13.35 & 26.72 & 20.04 & 17.78 & 21.43 & 26.73 \\
    Virgo & 37.95 & 64.02 & 44.36 & 28.98 & 36.59 & 50.36 & 10.63 & 38.63 & 37.27 & 21.16 & 41.67 & 33.07 & 33.18 & 53.88 & 37.45 \\
    \midrule \midrule
    Ours - SFT & \bfseries 50.77 & 71.83 & \bfseries 56.78 & \bfseries 50.35 & \bfseries 51.22 & \bfseries 62.68 & \bfseries 38.49 & 50.86 & 42.78 & 25.07 & \bfseries 64.25 & \bfseries 40.88 & \bfseries 50.65 & 60.78 & 44.16 \\
    Ours - SFT + DPO & 50.20 & \bfseries 73.08 & 55.26 & 45.94 & 48.29 & 62.09 & 37.98 & \bfseries 51.10 & \bfseries 45.41 & \bfseries 28.25 & 59.07 & 40.68 & 48.73 & \bfseries 62.28 & 44.70 \\
    \bottomrule
\end{tabular}
    }
    \caption{Results for all categories of the MMLU-Pro dataset.}
    \label{appendix:tbl:mmlu-pro-full}
\end{table}

% \textbf{High resolution evaluation.}
% \input{assets/appendix-table-high-res-docci-variants}

\section{Implementation Details}\label{appendix:sec:implementation_details}

\subsection{\longcot}

\textbf{Data generation.} Our framework consists of three stages: generates verifiable multiple-choice questions using $\LLM$, extracts simple chains of thought (CoTs) from vision-language models $\VLM$, and expands them into rich, long-form reasoning traces using frontier reasoning models $\RLLM$. 

\begin{enumerate}
    \item In Stage 1, we use \texttt{gpt-4o-mini-2024-07-18} with temperature $0.7$. First, we prompt GPT-4o using the prompt in Fig.~\ref{fig:prompt-description-to-mcqs} to generate multiple-choice questions. Then, we parse the outputs by prompting GPT-4o again using the prompt in Fig.~\ref{fig:prompt-parse-mcqs}.
    \item In Stage 2, we use Qwen2.5-VL-7B-Instruct with temperature $0.7$, top\_p $0.8$, repetition\_penalty, $1.05$, and set number of samples per input to 3
    \item In Stage 3, we use R1-Distill-Qwen-32B with temperature $0.7$, top\_p $0.8$, top\_k $50$, repetition\_penalty, $1.05$, and set number of samples per input to 3. To avoid outputs include phrases like ``As the description says'', we explicitly define \texttt{bad\_words} as ``describe, description, described, describes, descriptions, mention, mentions, mentioned, misread, text, stated, says, mental''
\end{enumerate}

\subsection{Training details}

\textbf{SFT Training.}
We fine-tune the language decoder with a batch size of $256$, sweeping learning rates over $\{10 ^{-5}, 8 \times 10^{-6}, 6 \times 10^{-6} \}$. Training runs for up to 5 epochs with early stopping based on the average validation accuracy. We set the maximum image resolution to $512 \times 512$ and the input cutoff length to $1024$.

\textbf{DPO Training.}
We fine-tune the language decoder with a batch size of $256$, sweeping learning rates over $\{1 \times 10^{-6}, 5 \times 10^{-7}, 1 \times 10^{-7}\}$. Training runs for up to 3 epochs with early stopping based on the average validation accuracy. We set the maximum image resolution to $512 \times 512$ and the input cutoff length to $1024$. For DPO, we set $\beta$ to $1.$ and following~\cite{pang2024iterative}, we include SFT loss with a weight of $0.5$.

\subsection{DOCCI Captions}
\label{subsec:docci_captions}
We select the same 500 images used to generate our dataset. Next, we format the training dataset with the user prompt "Provide a detailed description of the image.", prepending the image token and use the dense description provided in the dataset as the target answer of the model without further processing. We train the model using learning rate $8 \times 10^{-6}$ with batch size $256$ for a maximum of 20 epochs. The training reaches maximum average accuracy on the validation set in the third epoch and we subsequently use this checkpoint to report results in the main paper.

\subsection{VLAA-thinking}
\label{subsec:vl_thinking}

We preprocess the dataset into two different versions, discarding samples where no reasoning trace could be extracted. The first version uses $24,035$ randomly selected samples from the original dataset containing $158,827$ samples. The second version also $24,035$ random samples, however, we filter the dataset for images from ALLaVA-LAION and VizWiz. The latter specifically contains natural images - similar to the setting we train and evaluate on. We use the official dataset\footnote{\url{https://huggingface.co/datasets/UCSC-VLAA/VLAA-Thinking}} provided and apply some minor processing to the dataset to format the samples into a similar format as ours. In particular, we extract the thinking process and the answer from the \textit{ds\_answer} column of the dataset and place these into \texttt{<think>} and \texttt{<answer>} tags respectively. We use the same system prompt as for our model (see Sec.~\ref{appendix:sec:prompts}).

\textbf{Training.} We finetune the language decoder using batch size $256$. For both versions, we perform hyper parameter tuning by sweeping learning rates $\{10 ^{-5}, 8 \times 10^{-6}, 6 \times 10^{-6} \}$. We train for a maximum of 5 epochs and perform early stopping based on the average accuracy on the validation datasets. 

\subsection{Virgo}
\label{subsec:virgo}

We use the dataset introduced in Virgo~\citep{du2025virgopreliminaryexplorationreproducing} as $D_{SD}$\footnote{\url{https://huggingface.co/datasets/RUC-AIBOX/Virgo-Visual-Long-Thought-Dataset}} as other versions are not publicly available and it provides the best average performance in their experiments. As instructed on the webpage we use the "\texttt{conversation}" column of the dataset which the authors report to be the final data used to train the Virgo-7B model. The \texttt{conversation} column is constructed as the correct response with the shortest length of 5 samples given each prompt.

We apply minor processing to the dataset to follow our format by replacing the \texttt{<|begin\_of\_solution|>} and \texttt{<|end\_of\_solution|>} with \texttt{<answer>} and \texttt{</answer>}. Similarly, we replace \texttt{<|begin\_of\_thought|>} and \texttt{<|end\_of\_thought|>} with \texttt{<think>} and \texttt{</think>}. 
Finally, we append "Format the answer with the letter of the correct option in parentheses." to the system prompt if the question is a multiple choice question.  Overall, the resulting training dataset contains $14,540$ samples.

\textbf{Training.} 
For training, we follow the setup described in \ref{subsec:vl_thinking}, \ie, performing basic hyper parameter tuning, with the only change to limit training to 3 epochs as we found that the model performance peaks early during training. Surprisingly, we achieve the best validation performance before the first epoch ends.

\subsection{Evaluation}
\textbf{Inference setup.}
We use \texttt{vLLM}~\citep{vllm} for inferencing all models with greedy decoding. Detailed settings can be found in Tbl. \ref{tab:vllm_settings}. Further, we resize images' longer side to $512$ pixels preserving the aspect ratio if necessary.
As the reasoning traces for MMLU-Pro tend to be longer for all models due to the difficulty of the task, we double the maximum number of new tokens generated. 
We use four NVIDIA RTX6000.

\begin{table}[htp]
\centering
\begin{tabular}{@{}ll@{}}
\toprule
Setting          & Value \\ \midrule
cutoff\_length   & 2048 \\
max\_new\_tokens & 1024 (2048 for MMLU-Pro)  \\ 
temperature & 0.0  \\ 
top\_p & 1.0  \\ 
dtype & half \\
\bottomrule
\end{tabular}%
\caption{\texttt{vLLM} inference settings.}
\label{tab:vllm_settings}
\end{table}

\subsection{Training and Evaluation Prompts}\label{appendix:sec:prompts}
We provide the prompts for training and evaluation:

\begin{enumerate}
    \item Fig.~\ref{fig:prompt-train-description}: The prompt used to train VLMs on DOCCI descriptions.
    % \item Fig.~\ref{fig:prompt-train-mcqs}: The prompt used to train VLMs on DOCCI MCQs. \AL{Do we report DOCCI MCQs somewhere? We have the results of it except on MMLU-Pro}
    %\item Fig.~\ref{fig:prompt-train-vqas}: The prompt used to train VLMs on DOCCI VQAs.
    \item Fig.~\ref{fig:prompt-eval-answer}: The prompt used to evaluate VLMs to provide direct answers.
    \item Fig.~\ref{fig:prompt-eval-think-answer}: Inspired by the prompt provided by DeepSeek-R1~\citep{deepseekai2025deepseekr1incentivizingreasoningcapability}, we design the prompt used to evaluate VLMs to provide thoughts and answers. 
\end{enumerate}

\section{Qualitative dataset example}
\label{appendix:sec:qualitative_example}
\begin{figure*}[t]
    \centering   
    \includegraphics[width=0.5\textwidth]{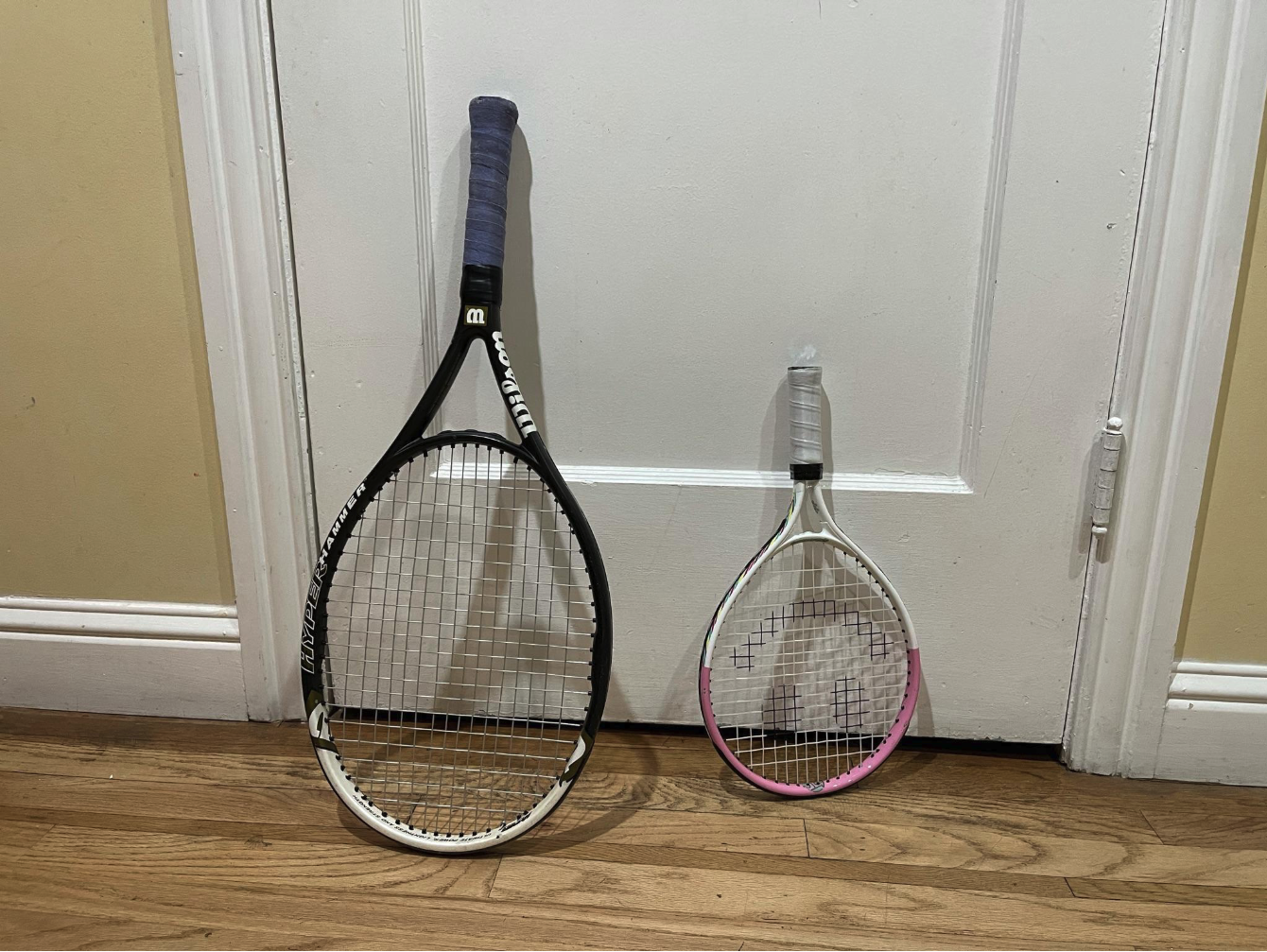}
\begin{tcolorbox}
\begin{Verbatim}[breaklines=true, commandchars=\\\{\}]
\textbf{Question}: How is the smaller tennis racket decorated within its strings?
Select from the following choices.
(A) A smiley face
(B) A floral design
(C) A logo
(D) A pattern of stars
% Correct answer: (A) A smiley face
\end{Verbatim}
\end{tcolorbox}

\begin{tcolorbox}
\begin{Verbatim}[breaklines=true, commandchars=\\\{\}]
\textbf{Simple CoT}
<think> 
Upon examining the smaller tennis racket, I can see that there is a distinct pattern of stars within its strings. This pattern is consistent across the strings, making it clear what the design is.
</think> 

<answer> (D) </answer>
\end{Verbatim}
\end{tcolorbox}

\begin{tcolorbox}
\begin{Verbatim}[breaklines=true, commandchars=\\\{\}]
\textbf{Ours}
<think> 
Upon examining the smaller tennis racket, I can see that there is a distinct pattern of stars within its strings. This pattern is consistent across the strings, making it clear what the design is. Wait, no, hold on. Actually, the image shows a smiley face inside the strings, but it's upside down. So that must be the decoration.

Wait, maybe I got confused. Let me double-check. The smaller racket has a smiley face printed in black within the strings. So the correct answer should be A smiley face. But wait, the options include both A and C. Hmm, could it be a logo? No, because the Wilson racket has a logo, but the smaller one has a smiley face. So the decoration is a smiley face, which is option A. 
</think> 

<answer> (A) </answer>
\end{Verbatim}
\end{tcolorbox}
    
\caption{Qualitative example of our \longcot dataset performing verification.}\label{fig:example}
\end{figure*}

We provide an example of our dataset in Fig.~\ref{fig:example}.

% Putting prompts at the end
\begin{figure*}[t]
\centering   
\begin{tcolorbox}
\begin{Verbatim}[breaklines=true, commandchars=\\\{\}]
System: You are an assistant that converts image descriptions to multi-choice visual questions.
User: Task:
You are given a detailed description of an image. Your goal is to generate diverse vision-centric, detailed questions that require a careful examination of the image for subtle visual details.  Each question should be answerable in a brief sentence or single phrase.
Instructions: 
- Focus on Visual Detail:
    - Ask questions that require examining fine details such as textures, patterns, and small or hidden elements.
    - Encourage the reader to analyze spatial relationships like object overlap, perspective, and layout.
    - Include aspects of lighting, shadows, and color gradients that affect the image's mood and depth.
- Comprehensive Coverage:
    - Ensure that the questions, as a group, address the majority of important details mentioned in the image description.
- Design for Multiple-Choice:
    - For each question, provide 4 answer options labeled A, B, C, and D.
    - Include one correct answer and three plausible distractors.
- Encourage Careful Inspection:
    - Design each question so that it cannot be answered without a close, careful visual inspection of the image.
    - Avoid generic or overly broad questions; each should target specific visual cues mentioned or implied in the description.
- Clarity, Specificity, and Brevity in Answers:
    - Formulate questions that are clear and focused on visual elements.
    - Ensure each question is detailed enough to challenge the reader to look beyond the surface.
    - Avoid questions that can be answered with general knowledge or assumptions.
    - Each question should be answerable in a brief sentence or even a single phrase.
- Structured Output:
    - Provide the questions in a numbered list.
    - Example Format: 1. <question> question here </question> <choices> (A) ... (B) ... (C) ... (D) ...  </choices> <answer> short answer here </answer>

Image Description:
\textcolor{red}{[IMAGE DESCRIPTIONS]}
Assistant:
\end{Verbatim}
\end{tcolorbox}
\caption{\textbf{Text prompt converting descriptions to multi-choices questions.}}\label{fig:prompt-description-to-mcqs}
\end{figure*}

\begin{figure*}[t]
\centering   
\begin{tcolorbox}
\begin{Verbatim}[breaklines=true, commandchars=\\\{\}]
User: You are given a text containing multiple multi-choice questions. Each question includes a question statement, several choices, and an answer. Your task is to reformat the text so that each multi-choice question follows the structure below:

1. <question> question text here </question> <choices> (A) choice A text (B) choice B text (C) choice C text (D) choice D text </choices> <answer> answer text here </answer>

Please ensure that:
- Each question is numbered sequentially (e.g., 1., 2., 3., …).
- The question portion is enclosed within the `<question>` tags.
- All answer options are enclosed within the `<choices>` tags, with each option clearly labeled (A), (B), (C), (D).
- The answer is enclosed within the `<answer>` tags.
- The original content is preserved, but any formatting issues are corrected according to the template above.

Here is the original content: \textcolor{red}{[PREVIOUS_RESPONSE]}
Assistant:
\end{Verbatim}
\end{tcolorbox}
\caption{\textbf{Text prompt to parse the response of Fig.~\ref{fig:prompt-description-to-mcqs} to multi-choices questions.}}\label{fig:prompt-parse-mcqs}
\end{figure*}

\begin{figure*}[t]
\centering   
\begin{tcolorbox}
\begin{Verbatim}[breaklines=true, commandchars=\\\{\}]
System: A conversation between User and Assistant. The user asks a visual question, and the Assistant solves it. The answer are enclosed within <answer> </answer> tags, respectively, i.e., <answer> answer here </answer>. Format the answer with the letter of the correct option in parentheses.
User: <image>Provide a detailed description of the image.
Assistant: \textcolor{red}{[IMAGE DESCRIPTION]}
\end{Verbatim}
\end{tcolorbox}
\caption{\textbf{Training prompt for training on DOCCI descriptions.}}\label{fig:prompt-train-description}
\end{figure*}

\begin{figure*}[t]
\centering   
\begin{tcolorbox}
\begin{Verbatim}[breaklines=true, commandchars=\\\{\}]
System: A conversation between User and Assistant. The user asks a visual question, and the Assistant solves it. The answer are enclosed within <answer> </answer> tags, respectively, i.e., <answer> answer here </answer>. Format the answer with the letter of the correct option in parentheses.
User: <image>\textcolor{red}{[QUESTION]}
Select from the following choices.
\textcolor{red}{[CHOICES]}
Assistant: 
\end{Verbatim}
\end{tcolorbox}
\caption{\textbf{Evaluation prompt for direct answers.}}\label{fig:prompt-eval-answer}
\end{figure*}

\begin{figure*}[t]
\centering   
\begin{tcolorbox}
\begin{Verbatim}[breaklines=true, commandchars=\\\{\}]
System: A conversation between User and Assistant. The user asks a visual question, and the Assistant solves it. The assistant first thinks about the reasoning process in the mind and then provides the user with the answer. The reasoning process and answer are enclosed within <think> </think> and <answer> </answer> tags, respectively, i.e., <think> reasoning process here </think> <answer> answer here </answer>. Format the answer with the letter of the correct option in parentheses.
User: <image>\textcolor{red}{[QUESTION]}
Select from the following choices.
\textcolor{red}{[CHOICES]}
Assistant: 
\end{Verbatim}
\end{tcolorbox}
\caption{\textbf{Evaluation prompt for thoughts and answers.}}\label{fig:prompt-eval-think-answer}
\end{figure*}

\end{document}